\documentclass[conference]{IEEEtran}
\usepackage{cite}
\usepackage[pdftex]{graphicx}
\usepackage[cmex10]{amsmath}
\usepackage{amssymb}

\usepackage{times}
\usepackage{array}
\usepackage[tight,footnotesize]{subfigure}
\usepackage[font=footnotesize]{subfig}
\usepackage{graphicx}
\usepackage{mathtools}
\usepackage{microtype}
\usepackage{amsmath}

\def\intersect{{\cap}}

\usepackage{alltt}
\usepackage{shadethm}
\newshadetheorem{comment}{Comment}
\def\ldirac{{\langle\kern -2pt\langle}}
\def\rdirac{{\rangle\kern -2pt\rangle}}
\def\Bldirac{{\Big\langle\kern -4pt\Big\langle}}
\def\Brdirac{{\Big\rangle\kern -4pt\Big\rangle}}

\def\boxx{{\vcenter{\vbox{\hrule height.3pt
          \hbox{\vrule width.3pt height6pt
          \kern6pt\vrule width.3pt}\hrule height.3pt}}\;}}

\def\impos{{\;\vcenter{\hbox{\rule{5mm}{0.2mm}}} \vcenter{\hbox{\rule{1.5mm}{1.5mm}}} \;}}

\def\lrarrow{\leftrightarrow \kern-8pt \rightarrow}

\def\2{\frac{1}{2}}

\def\i{\rm i}        

\def\beq{\begin{eqnarray}}
\def\eeq{\end{eqnarray}}
\def\2{\frac{1}{2}}
\def\AND{\,\cap\,}
\def\OR{\,\cup\,}
\def\with{~,~}

\newtheorem{example}{Example}

\newtheorem{definition}{Definition}

\def\lrarrow{\leftrightarrow \kern-8pt \rightarrow}

\def\frightarrow{\rightarrow \kern-11pt /~~}
\def\reducesto{\simeq \kern -3pt >}
\def\intersection{\cap}

\usepackage{fancyhdr}					
\begin{document}
\newcommand{\strust}[1]{\stackrel{\tau:#1}{\longrightarrow}}
\newcommand{\trust}[1]{\stackrel{#1}{{\rm\bf ~Trusts~}}}
\newcommand{\promise}[1]{\xrightarrow{#1}}
\newcommand{\revpromise}[1]{\xleftarrow{#1} }
\newcommand{\assoc}[1]{{\xrightharpoondown{#1}} }
\newcommand{\rassoc}[1]{{\xleftharpoondown{#1}} }
\newcommand{\imposition}[1]{\stackrel{#1}{\impos}}
\newcommand{\scopepromise}[2]{\xrightarrow[#2]{#1}}
\newcommand{\handshake}[1]{\xleftrightarrow{#1} \kern-8pt \xrightarrow{} }
\newcommand{\cpromise}[1]{\stackrel{#1}{\frightarrow}}
\newcommand{\policy}{\stackrel{P}{\equiv}}
\newcommand{\field}[1]{\mathbf{#1}}
\newcommand{\bundle}[1]{\stackrel{#1}{\Longrightarrow}}

\title{Quantitative Promise Theory\\\Large Intentionality and Inference in Autonomous Agents}

\author{Mark Burgess\\ChiTek-i AS\\May-June 2026}
\maketitle
\IEEEpeerreviewmaketitle

\renewcommand{\arraystretch}{1.4}

\begin{abstract}
  I discuss some quantitative representations of Promise Theory for
  processes involving autonomous agents.  Agent models are common in
  software systems, machine learning, and biology, for example, but
  may also apply to physics and other forms of engineering. I describe
  how Bayesian probability and information theoretic optimization,
  including Active Inference, may be incorporated with promise
  semantics---as well as how Promise Theory supplements solutions,
  helping to avoid probability's pitfalls, which include non-local
  coordination, calibrating, and normalizing probabilistic
  computations.  The role of boundary conditions in constraining
  allowed states and selecting decision thresholds is a form of
  promise, and agent alignment provides a scalable definition of
  intent.  Autonomous agents may congeal into swarms with superagent
  characteristics by trying to minimize their information, despite
  uncertainty that works to maximize it.  The use of Promise Theory
  involves some research challenges as well as stylistic preferences.
\end{abstract}

\hyphenation{similarity}
\tableofcontents

\section{Introduction} 

Promise Theory describes the behaviours and constraints of so-called
{\em autonomous agents}\cite{promisebook}. In Promise Theory, agents
are understood to be the focal dynamical element in all
phenomena. Think of them as somewhat analogous to the broad notion of
`particles' in physics. An agent is the embodiment of a standalone
process. A promise is a property or signal that announces an agent's
internal capabilities or intentions to other agents. Think of this as
a super generalization of the `charge' concept in physics.  In this
context, it may be of interest to infer, calculate, or predict the
evolution or trajectories of properties belonging to agents in space
and time.

Promise Theory was introduced\cite{burgessdsom2005} to clarify the
dynamics and the semantics of agent behaviour, stemming from their
behavioural constraints, especially in cases where this would normally
be taken for granted or would be presumed on the basis of symmetry
arguments. Promises are an idealization of intentionality, i.e. the
functional roles of agents within a `distributed system.  Agents are
a model of independent localized processes, which
`promise' to fulfill certain behavioural roles, and which may
therefore interact and form cooperative clusters to behave like
`superagents' on a larger scale.

Breaking with the traditions of multi-agent systems in computer science
\cite{agents}, promise theoretic agents are not assumed to be
centrally controlled machines; rather their behaviour is entirely
self-determined, and emerges through patterns of collective action at
scale.  Promise theoretic agents are thus causally independent, {\em
  active} elements in a process, rather than {\em passive} particles
driven by forces from without. One may call such elements `cognitive
agents' in the sense that they accept certain information from outside
and respond with a promised outcome, based entirely on interior
reasoning, which is the threshold hallmark for independent agency and cognition.

A promise plays a similar role to that of an equation of motion for a highly
localized, mostly isolated process. The interior evolution of agents
may be associated with the `keeping' of promised behaviours; the
exterior evolution is associated with interactions built on an
economics of trust.  This results in a fundamentally non-deterministic
picture of systems.

Much can be said simply by classifying constraints, and the bulk of
previous work has focused on revealing the limitations on process
semantics\cite{promisebook}, since this is where autonomy plays a key
role. It is tempting to use logical language for such constraints,
especially when applied to artificial systems. Nevertheless, there
remain operational questions of process rates and measures of
produced inputs and outputs, and so forth, which depend on promises.
The question I want to consider in this essay is what can be gained
from a quantitative analysis of agents?

Historically, the natural sciences have appealed directly to
continuous quantitative variables to capture causal responses, with
determinism implicitly expressed as a `known past' driving `possible
futures'. In natural science, established practices developed over
generations allow researchers to take many underlying assumptions for
granted, as a matter of long standing convention.  Their conversations
can therefore be abbreviated by a cocoon of implicit knowledge.
Promise Theory, by contrast, has a role in scenarios that are less well
established, and potentially less orderly. It adopts a `bottom up'
point of view, with such details exposed.

Where there is uncertainty, science often defers to statistics and
probability to smooth over the detailed issues with data sampling: this at the expense of an
intentional coarse graining with associated entropy. Such coarse
graining may result in a difference between `expected' and `intended'
outcomes of agents. As with quantum mechanics, a
partially observable constrained interior causality may result in
probabilistic external behaviour.

Agents are typically inscrutable processes, so one cannot know their
internal workings explicitly.  If we want to relate Promise Theory to
dynamical theories that involve probability, then it is natural to
begin by thinking of a promise as an autonomous constraint, i.e. a
context for local behaviour, which brings constraints to bear.  The
way a promise is expressed might, itself, be an imperfect
representation: coarse grained, approximate, with unknown dynamic
variables, etc, but it remains as the residual `intention' behind the
behaviours encapsulated by an agent.  What we observe is a form of
conditioned behaviour, so we explore effective descriptions along a
pipeline of mutual interactions, from promised origins to statistical
outcomes.  Both Promise Theory and Probability Theory have
conditionals, referring to different aspects of prediction, so we need
to be clear about their distinctions.

\subsection{Dynamical variables, representations, and  narratives}

What quantitative measures might we look for, or calculate, in a theory of agents? 
Some possibilities come to mind:

\begin{itemize}

\item Representative values of key dynamical variables associated with the
  behaviour of an agent, including exterior values that are promised and
  dependent on interior `hidden' variables.
  
\item An operational model of an agent's interior processes involved in `keeping
  promises', which includes generating output or handling input, including
  explaining finite resources and interior constraints\footnote{This
    is analogous to solving the physics of a mechanical system in a box, a potential
    well, or a central potential.}.
  
\item A collaborative model of exterior promise keeping, explaining
  the substance of the promise and how it is represented, how it
  benefits other agents, including sampling rates and
  transmission currents.

\item A model of agent learning, pattern recognition, and inferential decision making.

  \item Maps of the possible pathways or dynamical trajectories through
  an information pipeline.
  
\item Inferring an agent's intent, in the absence of a promise, by observing it over time.
    
\item Distributed reasoning, e.g. reasoning networks, swarm intelligence, etc.

\item Long range order: calculating the alignment of agents with a particular intention
 as a common property within a larger population, e.g. the agents
  infected with a certain promise or belief.
\end{itemize}
We must also consider the role of {\em coarse graining} and {\em resolution} of variables
in preserving information from discrete
samples.  What is the minimum variable resolution to support a given
process, e.g. long range wave transmission between agents?

The frameworks I'll consider are the following:

\begin{itemize}
\item {\em Promise Theory}: emphasizes autonomous processes,
  i.e. phenomena with strictly local variables, centered on cellular,
  bounded regions, linked with a graph topology.  It defines the
  protocols and limitations for interaction that result from strict
  accounting for causal independence.
\item {\em Probability Theory}: is a mixed picture that expresses
  methods for counting elements of sets. It does not depend on any
  particular scope, topology, or model of a system change.  It is a
  key tool in statistics and stochastic systems.
\item {\em Statistical and Quantum Mechanics}: are formalisms for
  analyzing physical scenarios, with mixed conceptual content.
  Statistical physics refers to a scale of bulk
  `thermodynamic' scenarios, while Quantum Mechanics refers to
  scenarios on the level of individual `particles'.
\end{itemize}
Probabilistic ideas further arise with three distinct interpretations:
\begin{itemize}
\item Past event frequencies (based on sufficient data about prior
  recorded behaviour, which is assumed to be accurate),
\item Bayesian beliefs (which concerns updating
    of prior estimates, which may or may not be accurate), and
\item Propensities (which are theoretical predictions of likelihood, not
based on data).
\end{itemize}
A propensity is a calculated value that resembles a {\em potential}
for future happening\footnote{Propensities were applied to statistical
  physics, including quantum mechanics by Karl Popper in
  1957\cite{popper}. This is the only plausible interpretation of
  probability for quantum mechanics.}.  Bayesian probability is
typically used for inference of propositions (hypothesis testing)
based on obviously incomplete data. Many authors do not specify which
of these semantics are being assumed.

\subsection{Shared reference concepts}

Every phenomenological model builds on a concept of space and time as
a representation of change, i.e. the theatre in which phenomena
unfold. The Euclidean space we use to model human level events is
largely irrelevant to what happens for agent systems. The space of
their configurations refers to interior variables, which
are bounded and separate from one another. Channels for communicating
influence describe the effective topology of an agent system, and the passing
of time is measured by what happens within each agent independently.

Every process is further represented with respect to its {\em
  invariants}.  There is an implicit assumption that what one collects
data about is a set of invariant, time-independent phenomena. So a
complete characterization of {\em repeatable} phenomena can be represented
by probability mass or density functions over these invariant characteristics. 
This is what leads to a consistent and thus meaningful representation, e.g.
for the classification of
events into different categories. The clearest way to define
probabilities is in terms of sets and subsets, so the invariances
typically represent a constant basis of sets that span a complete set
of possible outcomes.

We can define a number of terms here in order to establish a consistent language
and notation across descriptions.

\begin{itemize}
\item {\em Observables $a$}: Identifiable process characteristics, with
  distinct interpretational semantics. Observables represent the
  coherent concepts associated with process measurements (rather than
  their values), including ideas like energy, momentum, charge,
  potentials, signals, messages, etc.  Observables are derived from
  the values of dynamical variables, and many authors will use these
  concepts interchangeably.

\item {\em Dynamic variables}: Dynamical variables are the memory
  cells of a dynamic system, describing its capacity for change. These may also
  called its {\em degrees of freedom}. These are the quantities that change
  as a result of its evolution. Their values are what we try to
  predict in a theory, and what we interpret as the result.
  
\item {\em Measurement (event)} $e_a, a = 1,\ldots,M$: The extraction of a value for an
  observable. A particular observable may vary over a fixed set of
  possible values, either discrete or continuous, corresponding to
  different states. The action of sampling the variables to extract
  that value leads to a measurement.  The continuum hypothesis
  suggests that continuous variables are in fact approximations for a
  high density of possible discrete values.
  
\item {\em States} $\omega_i, i=1,\ldots,D$: Distinguishable configurations of dynamical
  variables that represent distinct attributes of a process---sometimes
  called `state of the world' as an indication that these are
  representative of a world model for the phenomena.  States
  contain the information of which is sampled by interacting with the
  system, thus they ultimately express the possible alternatives for a
  process change---and may characterize different qualitative
  phenomena. When we denote a state, we refer to a snapshot of the
  physical system as a vector of `simultaneous' dynamical variables.
  States may be composed over different scales: microstates and
  macrostates refer to snapshots at opposite ends of the spacetime
  scale.
  
\item {\em Constraints}: limitations on the ability of process
  variables to express certain values. These may be maintained
  autonomously or `voluntarily', or they may be effectively `imposed' by virtue of
  interacting with the environment. Constraints include promises and boundary
  conditions.
\end{itemize}

It's not uncommon to speak of states and their values interchangeably,
but I will try to be more precise. In a vector space, one may think of
a low level `micro' state as a vector, and its value as its
coordinates relative to a spanning set of basis vectors. In a
computational view, such states are vectors of memory.
The definition of `state' also varies considerably between different
phenomena and their scales. It is associated with a distinct configuration
of the degrees of freedom characterizing a system. Theoretically, such
variables may be speculative or merely an effective handle, introduced
for convenience, as we grapple with the unknown.

\subsection{Causality and determinism}

Prediction depends on how one state of a theory leads to another.
All theories have to deal with questions about causality and
determinism sooner rather than later. There is a certain `stylistic'
arbitrariness to dealing with what one doesn't know.

For example, a classical view of particles of matter is that they
spontaneously follow unique trajectories by riding a guiding
potential, supported by postulated conservation laws. They may
exchange momenta by `collision', etc. Measured outcomes are associated
with defining the end of a continuous trajectory, under the assumption
that such trajectories can be freely observed.
What is interesting here is the
separation of a description into two parts: an ambient guiding field, and a localized
particle: continuum and discreteness on display simultaneously.

In Quantum Mechanics, we replace classical trajectories with a
sequence of state transitions. There is a smooth trajectory for a
guiding field (the wavefunction), but a discontinuous selection of
subsequent state.  The theory predicts that only specific allowed
values can be measured, one at a time. The set of outcomes is
determined by the wavefunction which is determined by boundary
conditions, something like a promise; however, the selection of which
precise state is not determined, and is thus treated as probabilistic.  Nothing
in the theory explains how measurement outcomes are actually derived
from the theoretical evolution of probabilities---because we simply
don't know. We do know that these transitions cannot be observed, and
that observation has a significant impact on the unknown process. Thus, it is
not causal or deterministic on the scale of a single measurement, and
we work around the unknown by referring to propensity.

In Promise Theory, the presumed dynamical subjects are now {\em agents},
i.e.  encapsulated processes that are localized as a single entity\footnote{In fact, much of quantum
  mechanics can also be represented this way.}.  A `free' agent is
entirely self-determined. In the absence of influences promised by other
agents AND explicitly accepted by itself, an agent's own promised behaviour
is independent of external influence.  This reverses the conventional
notion that an upstream action causes a downstream outcome, because it leaves
the result to the selection of the downstream agent (known as the Downstream Principle\cite{promisebook}).

A hard line physicist would perhaps see no contradiction in arguing
that that nature has no choice but to obey mathematical rules from
outside, while explaining that the selection of a particular outcome
is an unknown choice, thus elevating probability and non-locality to
something fundamental. In a Promise Theory view, such a choice would
be explained by saying that the result of unobservable processes
happens within agents, allowing the character of probability to remain
unchanged.

Of Quantum Mechanics, 
Schwinger wrote\cite{schwingerQM}: ``By {\em causal} we mean
... inference in time. Given the state of affairs at one time, the
state at another is uniquely determined. What makes it deterministic
is that knowledge of the state also determines all phenomena
precisely.'' Then later, he refers to the dualism of
interpretation, with both a wavelike (field)
nature and a particle nature to match the limiting classical
representation of change: ``The field supplies the dynamical agency by
which the particle interacts'', which returns to the classical approach.

The concept of a state as an internal configuration of agents allows
us to assign a probability to that configuration in a repeatable way,
just as pure quantum states assign probabilities.  A state, in the
statistical sense, is a specification of a probability distribution
with each observable.

The picture of time itself is important to these definitions, and it
generally taken for granted. Observationally, each sampling of the
system's state takes a finite amount of time, as specified by the
Nyquist-Shannon sampling theorem. Thus every observation and
characterization of a dynamical system corresponds to a coarse grained
approximation of the system.  The lifetime or persistence of states
depends on their scale and is matter of both definition and
measurement.  While a `macro' or statistical state is a realizable exterior
configuration, a single eigenstate from a complete set
of spanning functions is an interior configuration.


\section{Promise Theory}

From a dynamical systems perspective, a promise is a constraint on an
agent's behaviour.  Promise Theory deals with the properties of
autonomous agents, i.e. active entities which are, a priori, causally
independent of one another.  Autonomy is a strong constraint on what
is causally possible for agents, since all changes must originate from
within the agent until is chooses to accept outside influences. The
progressive relaxation of autonomy to allow interaction shapes the
behaviours of initially autonomous agents.
The fundamental tenet of Promise Theory is that:
{\em ``No agent may promise
  anything on behalf of any agent but itself.''}.
In other words, the last word about all change originates from within the agent.
Either an agent decides for itself, or it can veto suggestions from outside.
The concept of `agency'
is associated with this boundary for influence. One should note that
such autonomous agency is a stronger constraint than the Markov condition,
which already assumes that events can influence one another.

Promise Theory concerns, in large part, how initially autonomous
agents voluntarily forego their autonomy (without giving it up
completely) to cooperate in some process on a scale larger than a
single agent.  This is sometimes called the formation of a superagent.

A promise is a signal concerning the `intent' of an agent to constrain
its behaviour in some way.
\bigskip
\begin{definition}[Intent]
  A form of persistent and coherent behavioural bias, whose residual
  signal is reflected in a stable probability distribution for times that are long
  compared to observational events.
\end{definition}
\bigskip
Notice that---in this view---`intent' is not a term reserved for humans or higher animals. It is associated
with selection of dynamical variables that guide and shape behaviour.
By signalling an intention to another agent, an agent is said to make a promise.
A promise is neither a guarantee, nor an obligation. It is merely an autonomous declaration
of an intention that may or may not be accurate.

\subsection{Notation for promises}

The generic label for agents in Promise Theory is $A_i$, where Latin
subscripts $i,j,k,\ldots$ numbers distinguishable agents for
convenience (these effectively become coordinates for the agents).  We shall often
use the symbols $S_i$ and $R_j$, instead, for agents to emphasize
their roles as source (initiator) and receiver (reactor).
So the schematic flow of reasoning is:
\begin{enumerate}
\item $S$ offers (+ promises) data.
\item $R$ accepts (- promises) or rejects the data, either in full or in part.
\item $R$ observes and forms an assessment $\alpha_R(.)$ of what it receives.
  An agent assesses all information it receives using a variety of functions
  for different assessments, depending on its capabilities.
\end{enumerate}
Observations and assessments are limited only by the capabilities of
the agent.  They may be economic assessments, to minimize or maximize
some parameter. They may be semantic assessments to stay close to a
logical or symbolic constraint. Assessments are a centre-point for the
indeterminism of agent systems.  One should be careful in
anthropomorphizing agency too much. We use words like promise, assessment,
etc because the convey the correct meaning in the limiting case of
complex human agents, but equally assumptions have to be scaled back
when being used for simpler agents with limited capabilities. For a
molecule, an acceptance of a promise may be equivalent to accepting a
donor molecule by docking with it.

The elementary agents of a system embody {\em processes}, any of which
may express promises about state and services. Processes are hosted at agent locations 
$A_i,S_i,R_i$ etc.
Agents have finite resources, so their capabilities are limited. However, they might
over-promise accidentally or deliberately, e.g. promise 10 parking spaces for 20 cars,
or 100 seats for 150 passengers. Such a promise can be resolved by time-sharing.

\begin{figure}[ht]
\begin{center}
\includegraphics[width=5cm]{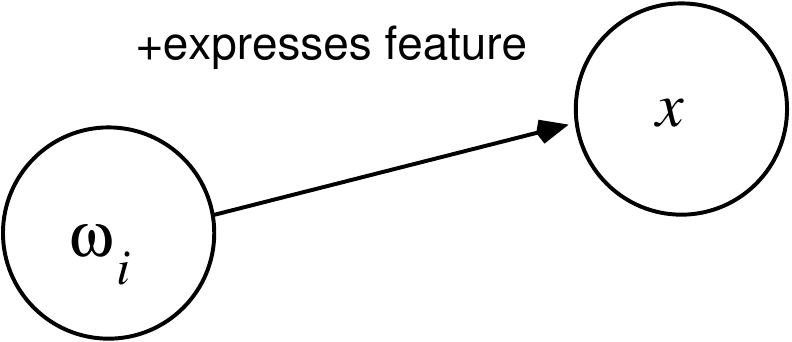}
\caption{\small Notation for promise graphs and semantics is
  fundamentally graphical and (despite a few overlapping concepts) is
  not equivalent to that for Bayesian networks.\label{notation}}
\end{center}
\end{figure}

\subsection{Promise parameterization}

Promise bodies affect only the agent making the promise, but may refer to its intentions towards
other agents, e.g.
\begin{itemize}
\item A scalar promise refers to no other agents, e.g. I promise to brush my teeth $+b$.
\item A vector promise refers to a single third party, e.g. I promise to follow orders from $A_3$: $=b(A_3)$.
\item A tensor promise may refer to any number of agents, e.g. I promise to get the best deal from
suppliers $+b(A_1,A_2,\ldots,A_n)$.
\end{itemize}
Promise also come in two major flavours.
An {\em offer} promise, written with body $+b_i$ made by $S_i$ to
$R_j$ is written:
\beq
A_i\promise{+b_i} A_j,
\eeq
where the $+$ refers to an offer of some information or behaviour (e.g. a service).
This is a part of $A_i$'s autonomous behaviour, and the promise constrains only $A_i$.

$A_j$ may or may not accept this offer, by making a dual {\em acceptance} promise, marked $-b$ to denote the
orientation of intent:
\beq
A_j\promise{-b_j} A_i.
\eeq
Only if {\em both} offer and acceptance promises are kept can an influence be expected to
pass from $A_i$ to $A_j$. In general, the offer and acceptance may not match
precisely, in which case the propagated information will be the overlap (mutual information)
\beq
b_{ij} = b_i \intersection b_j,
\eeq
in the manner of mutual information\cite{shannon1,cover1}. There is nothing
probabilistic yet: this is all assumes to be intentional and attended by best effort.

\begin{figure*}[ht]
\begin{center}
\includegraphics[width=15cm]{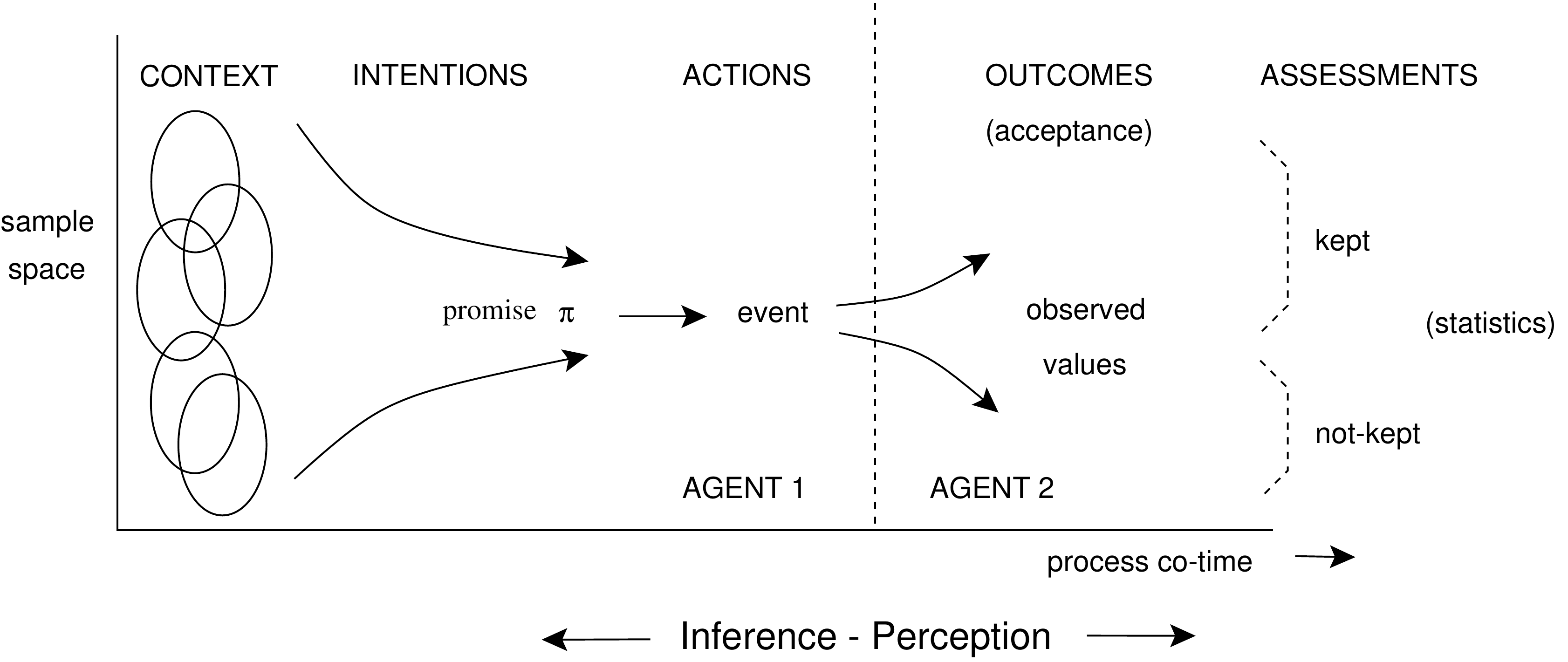}
\caption{\small The `data pipeline' in a promised causal influence.
  An agent promises its own behaviour, i.e. a subset of possible
  outcomes of its own behaviour. This constrains events associated with the
  promised behaviour. Another agent, accepting this promise may observe the outcomes, measure 
  and count them, assess whether they keep the promise or not.\label{promise}}
\end{center}
\end{figure*}

What form does the abstract $\pm b$ take? A form of signalling between
them may take the form of a language, from regular language all the way up to
advanced (human) language\cite{lewis1}.

The result of a promise is the assessment of an outcome:
\bigskip
\begin{definition}[Outcome]
  A distinct event providing a value for a dynamic variable constrained by the
  promiser.  Outcomes form a set, like a menu of possible results, which may be either discrete or
  continuous.
\end{definition}
\bigskip

\subsection{Conditional promises}

Agents can make promises conditionally.
To promise $b$ on receipt of a certain conditional $c$ (e.g. if the book arrives, I will read it by Monday):
\beq
A_i \promise{+b | c} A_j.
\eeq
This notation is distinct from that used in conditional probabilities, which
has a distinct meaning. Such a conditional offer 
is not yet a promise. It must be completed by also promising information
about the condition. To complete a conditional promise, an agent must
also promise the condition to be true (to best effort),
i.e. the offer of some conditional $c$ from another upstream agent $A_C$,
\beq
A_C \promise{+c} A_i
\eeq
must be taken together with the downstream acceptance of $c$, and
since this is purely local, it the downstream must signal its acceptance to $A_j$
and then confirm this even farther downstream to complete its intention
to keep the promise\cite{promisebook}:
\beq
A_i &\promise{-c}& A_C\nonumber\\
A_i &\promise{-c}& A_j
\eeq

Such conditionality is subtle and potentially misleading, as it
involves agents as `middle men', which introduces non-deterministic
aspects. Agents are not reliable relays in the sense one might expect.

A promise that is made subject to a condition is not a complete
promise, unless the state of the condition that predicates it has also
been promised. It has incomplete information. Such a promise is hence
null-potent. 

If a promise with body $S$ is provided subject to the provision of
a pre-requisite promise of body $b$, then the provision of the pre-requisite by
the same agent is completely equivalent to the unconditional promise being made:
\beq
A_1 \promise{b|c} A_2 \with A_1 \promise{+c} A_2  \equiv
A_1 \promise{b} A_2
\eeq

This gives us a rewriting rule for promises made by a single agent in the promise graph.
The $+$ is used to emphasize that $X$ is being offered, and to contrast this with the next
case.

Consider now the first case in which one agent assists another in
keeping a promise. According to our axioms, an agent can only promise
its own behaviour, not that of other agents, thus the promise
basically comes from the principal agent, not the assistant, which
might not even be known to the final promisee\footnote{When a
conditional promise is made and quenched by an assistant, the
`contact' agent is directly responsible by default. We shall refine
this view with alternative semantics later, since this is all a matter
of managing the uncertainty of the promise being kept. As soon as we
allow rewriting rules of this basic type, it is possible to support
multiple solutions for bringing certainty with graded levels of
complexity.}.  Assistance is a matter of voluntary cooperation
on the part of the tertiary agent.

There is second case, in which an agent signals its intent to
rely on another.
If a promise with body $b$ is provided subject to the provision of
a pre-requisite promise $c$ by another agent, then the provision of the pre-requisite by
an assistant is acceptable if and only if the principal promiser also promises
to acquire the prerequisite $-c$ from an assistant (promise labelled $-c$):
\beq
\left.\begin{array}{c}
A_1 \promise{b|c} A_2 \\
A_1 \promise{-c} A_2
\end{array}
\right\rbrace
\sim A_1 \promise{b} A_2
\eeq

In neither case does a conditional promise, or indeed any promise conditional
or otherwise, have a certain outcome.
It only signals its intent to aim for a best effort outcome. The dynamical
uncertainties remain in either case, but may be greater in the latter case
of accepting help from a second agent.

\subsection{Assessment}

\bigskip
\begin{definition}[Assessment]
An {\em assessment} is an agent's autonomous act of
classifying the outcome of an event with respect to the promises
it has received. An assessment can be made by any agent in scope of the promise:
both the promiser/sender and the promisee/receiver, as well as third parties, may
observe what is happening and make such a determination.
\end{definition}
\bigskip An assessment may simply fall into the category of `promise
kept' and `not kept', or may be classified into a menu (distribution)
of mutually exclusive states $\omega_i$ that we may refer to as the
`pure states' of the promiser.

When we write an assessment $\alpha_a(\pi)$ for some agent, the
subscript is a reminder that the assessment is made on $A$'s terms,
according to $A$'s information, on a schedule run by $A$'s clock.
Thus it refers to $A$'s local rate of time and is subject to all
possible distortions experienced (either knowingly or unknowingly) by
$A$. This is why we must invoke the downstream principle in all
measurements, including those that are the end results of some chain
of interactions.  Local clock rates may affect local assessments of
reliability, errors, or be absorbed into the `noise' of the system.

At this stage, everything about the symbolic descriptions refers only to the
topology of agents and the semantics of their intentions. This is `protocol'. Any dynamics
are merely implicit: there is no explicit transactional process or quantitative
characterization of the agents in time.
The next phase is to describe agent actions.

\subsection{Spacetime embedding}

The graph of promises  between agents forms the
spacetime on which behaviour is played out. How agents move or change with
respect to one another is an additional layer of description.
\begin{itemize}
\item If agents are embedded, like cells in a biological environment, one
may add motion to the agents in some embedding space (called motion of
the first kind)\cite{spacetime1,spacetime2,spacetime3}.

\item If agents' relative positions are fixed or are outside the realm
  of description, like a solid state, and we only care about what they
  promise to one another, one may still characterize the flow of
  information from one agent to another (motion of the second and
  third kinds)\cite{virtualmotion1,virtualmotion2,virtualmotion3},
  e.g. supply chains through ports, waves passing through a crystal or
  a neural network.
\end{itemize}
The advancement of time in a collection of agents suffers from the same complications
as relativity in physics. One can choose to adopt a universal time for all agents,
according to some outside observer's clock (the Galilean model). However, autonomous agents
have `private scope' (in computer science terminology) or are `local' (in physics terminology).
Thus, what one agent experiences as time is based on its own interior dynamics and
exterior time is only observed through that lens. This complicates causal determinism in
a similar way to the Lorentz-Poincar\'e relativity formulated by Einstein\cite{virtualmotion1}.

Probability is an appropriate tool to discuss the quantitative aspects
of agent promises: there is an implicit indeterminism to agent
interactions, which can be absorbed by a probabilistic description.
So to realize a probabilistic description for agents, we must
determine what events a promise theoretic system is counting and under
what conditions.

\subsection{Observation and measurement}

Autonomous agents are, by definition, causally independent systems.
They observe one another's promised outcomes, by a process of
sampling, based on their own independent clock and resources.  By the
Nyquist-Shannon sampling law\cite{shannon1,cover1}, any such process is
necessarily non-deterministic, unless every agent can be twice as fast
as the one before it (which would be impossible in general). Thus,
what one agent promises may not be perceived exactly as intended by
the recipient or promisee.

When characterizing a probabilistic system, one must, in addition to
characterizing a distribution of possible outcomes, say something
about the expected invariances of the system. The relying on data
alone involves a certain risk that circumstances may, in fact, have
changed.

Under steady state conditions, one implicitly assumes that what
has occurred in the past may be a reliable guide to what may happen
again, in the future---which is by no means given. 

\subsection{Scaling of agency and clocks}

A universal characteristic of autonomous agents is that they operate
according to their own internal resources and their own internal clocks. The rate at
which we observe processes in other agents involves {\em relativity}
of their individual rates of change.  In general we don't know `how
fast' agents react to process events, and we cannot necessarily assume
homogeneity amongst agents or their response rates, so there is not
always much we can say about the processing rate of autonomous collectives.

However, regardless of their individual rates, we can say the
following: as several agents and their promises are composed to form a
collaborative process (effectively making several agents promise
something as a single scaled `superagent'), the resulting `collective'
promises they make, at their effective superagent boundary, can only
be delivered at a rate that is slower than that of any of the individual agents.

If an observer $A_i$ samples
the results of a promise every $\Delta t_i$ time units, then it can never
observe any change that happens faster than this. An agent with a finite
sampling rate will experience `jumps' in promise outcomes when observing
an upstream process that might be smooth according to a faster agent.
There are several timescales in competition. Suppose we try to measure
all agents according to some universal time, faster than all other agents $\Delta t$.
\begin{itemize}
\item The rate at which $S$ keeps its promise $\pi_S^{(+)}$: ticks every $\Delta t_S$ universal units.
\item The rate a which $R$ keeps its promise $\pi_R^{(-)}$ to accept
  information: ticks every $\Delta t_R$ universal units.
\item The time it takes for $R$ to assess the outcome of the promise binding: $\alpha_R(\pi_R^{(-)}\pi_S^{(+)})$.
\end{itemize}

Now consider a cluster of agents that collaborate in promising a result that is
obtained from a series of conditional promises:
\beq
\pi_1(x) \mapsto \pi_2(x'|x) \mapsto \pi_3(x''|x') \mapsto \ldots
\eeq
If $\pi_i$ takes $\Delta t_i$ universal time units to complete, then $R$'s assessment of
the 
time for the chain to complete must lies somewhere in the range
\beq
 \max_i \Delta t_i \le \alpha_R(\Delta t_{1+2+3+\ldots}) \le \sum_i \Delta t_i,
\eeq
where this range depends on the resolution of the observer's time. To this, we must add
the time to receive and assess as additional uncertainties.

We are used to building clocks with simple mechanisms that are fast. However,
when we reach atomic scales, all processes occur on a comparable timescale and
measurements become less reliable.
Measuring the completion time of agents is a basic characteristic for
agent systems. When measuring times, we must always specify which agent is measuring the interval,
since every autonomous agent has its own clock, running at a rate that depends
on its own capabilities and available resources.

\subsection{Instantaneous and average variables}\label{avgvar}

An agent's behaviour is based on interior variables. These are usually unobservable
to outside agents unless their values are promised explicitly. Suppose an agent has a set of
$\Omega$ such variables that characterize its state. Then any promisable quantity
can be represented by some variable $x_i$:
\beq
x_i = \sum_{a=1}^D \; w_{ia}\,\omega_a,
\eeq
where $w_{ia}$ is a set of weights and $\omega_a$ is a basis set of
vector `features' that we may call the pure states of the agent.
Agents may have additional memory in addition to that which determines its
behaviour.

Memory may store semantic details, e.g. proper names, acceptable client list such as
proteins that the agent allows to dock with receptors, or access control on an
E-commerce platform, etc. 

Agents can learn quantitative average values by sampling data from
inside or outside.  Moving average values, e.g. the number of clients
being serviced at any one time.  Obviously an agent cannot collect
`big data' to analyze. It has to make do with finite memory. There are
two ways of maintaining running averages with a fixed memory allocation.
Both require agents to be able to represent real numbers internally.

\begin{enumerate}
\item {\em A fixed window for events:} Suppose we wish to keep
  a window of $N$ past events. The agent learns data $x_1, x_2, \ldots, x_N$,
  where the order is maintained. The average is computed as
  \beq
  \overline x^{(N)}(t) = \frac{1}{N}\;\sum_{i=1}^N\; x_i.
  \eeq
  On each update, the first value (divided by the normalizer $N$) is dropped and replace
  ed by a new:
  \beq
\overline x^{(N)}(t+1) = \overline x^{(N)}(t) - \frac{x_1}{N} + \frac{x_{N+1}}{N}.
\eeq
This approach has several disadvantages, especially for
elementary agents. It requires a memory size of $N$ values and the ability to
perform addition, subtraction, and division! These are not elementary capabilities. 

\bigskip
\item {\em Controlled forgetting:} The alternative is to use a
  geometric updating procedure that only uses addition\cite{burgessDSOM2002,burgessC14}.
  If we define a new expectation operation by this recurrence relation:
  \beq
\langle x\rangle_{t+1} = \lambda \,\langle x_t\rangle + (1-\lambda)\, x_{t+1},\label{runav}
\eeq
where $\lambda \le 1$, then we now have an average with a policy for forgetting. As this expression is
iterated $N$ times, the oldest values accumulate factors of $\lambda^N$ and thus
become geometrically less important to the average. Apart from being more computationally
efficient, this only uses a single memory representation and can tune its
dependence on new information versus old. The variance of the window is simple
to calculate in the same way, since the variance is just the average of the square
difference:
\beq
\sigma^2_{t+1} = \langle (x-\langle x\rangle_t)^2 \rangle_{t+1}.
\eeq
Thus, even an elementary agent, with little memory, can easily calculate the
average and variance of data (though not necessarily the standard deviation, which involves
more difficult computation).
\end{enumerate}
\bigskip
\begin{example}[Monitoring software agents]
  An agent measures a set of features about its runtime environment: $\vec x$,
  whose components are each updated according to eqn (\ref{runav}) with some
  policy value $\lambda_i$, in general. Typical features are quantities like:
  percentage CPU usage, memory usage, input/output transfer level, number of
  clients, etc\cite{burgessC14}.
\end{example}
\bigskip
\begin{example}[Sensory data aggregator]
  An agent promises to receive data from $N$ upstream sensors to form
  an image $\vec x$ in a set of discrete samples separated by
  dead-time $\Delta t$. In order to smooth a persistent image, given that
  the exact arrival times are unknown, it applies eqn (\ref{runav}) to
  accumulate running averages that change significantly with new data by
  choosing a high values of $\lambda$ for all the sensors. In this way
  is balances past experience and new experience without risking an empty
  sensor measurement as a result of poor timing etc.
\end{example}

\bigskip
\subsection{Thresholds and decision making}

The defining property of an autonomous agent is its ability to reject
outside influence and determine its own trajectory. When we attempt to
calculate the behaviour of an agent, we have to determine the
assessment functions $\alpha_A(\pi)$ of the agent $A$, and the
promises it makes and accepts.

Elementary agents, i.e. those with few
interior degrees of freedom will have little capacity to make
decisions to accept or reject influence based on complex criteria, so
we can assume they will tend to accept inputs uniformly without modification, to
the best of their capabilities; but larger agents such as animals and
humans with many interior resources will make much less predictable
choices. Even low level agents must satisfy the Nyquist-Shannon sampling
theorem, and be subject to limitations associated with finite resources,
so it would be wrong to assume that all machine-level communication is
deterministic and without error. Many agents will have developed error
correction mechanisms that we take for granted.

In reasoning systems, i.e. systems that can discriminate between different
classes of behaviour, arbitrary thresholds between the classes must inevitably
be used to separate discrete outcomes. There is a finite set of states
$\Omega = \{\omega_i\}$ into which a process naturally classifies a decision.
\begin{itemize}
\item The phenomenon may be sufficiently constrained to make these choices clear, e.g.
when geometry constrains activity and leads to a discrete set of eigenstates, as in
quantum mechanics, or animal varieties based on digital genetics. 

\item An agent's finite resources may lead to an arbitrary cut-off, or coarse-graining
  procedure for value ranges, yielding a set of arbitrary bands, like radio station frequency
  ranges.
\end{itemize}
To assess behaviour, we need to know how agents' assessment functions $\alpha_A(\pi)$ are defined.
\bigskip
\begin{example}[Wave propagation through agents]
  Consider a simple elementary case: a finite linear chain of agents $A_i$, which promise to inform their neighbours
  of a real valued displacement function $\psi(A_i)$, where $1 \le i \le L$.
  The sharing of interior variables is promised as follows:
  \beq
  A_i &\promise{+\psi(A_i)}& A_{i\pm 1}\\
  A_{i\pm 1} &\promise{-\psi{A_i}}& A_i\\
  \alpha_{i\pm 1}(\pi_i) &\mapsto& \psi_i
  \eeq
  and the agent's interior response to its neighbours is promised by:
  \beq
  &A_i&\promise{\psi \mapsto \psi + \Delta \psi} A_i\\
  &\Delta \psi& = \frac{(\psi(A_{i\pm 1}) - \psi(A_i) )}{m}
  \eeq
  where $m \ll 1$. Notice that the symmetrical promising to agents on either side $i \pm 1$ means that
  the promise graph is an undirected, but totally ordered sequence.
  The virtual field $\psi(A_i)$ is like a string
  which can transmit displacements along its length. If we start a simulation with a string plucked at
  its centre, with $\psi_{L/2}$, then the displacement will oscillate away from the centre and spread
  out, interfering as it reflects back---something like simple harmonic motion.

  We should note that agents do not obviously obey a conservation of energy as
  one might expect in physical simple harmonic motion. If the balance of
  values is maintained precisely, this might be approximated.
  Indeed, it is
  extremely difficult to build any kind of conservation laws into agent
  motion\cite{virtualmotion3}. That is because, by Noether's theorem,
  conservation laws are proxies for the smooth continuity of
  spacetime. In a discrete system, there is no such continuity, even
  if agents are identical. The system is not really in a dynamic equilibrium;
  rather it is in an approximate non-equilibrium steady state.

  Harmonic motion also requires a continuous
  displacement function, which requires a small increments only, i.e. a small
  value for $m$; a discrete set of values will not result in
  anything like wave motion. A small loss of accuracy causes the
  motion to overshoot its maximum limit and increase to infinity
  rather than turn around and come back in the opposite direction.

  Notice also that there is no need for any normalization of the $\psi$
  interior variables. 
\end{example}
\bigskip
In this example, the spacetime of the agents is totally ordered with no loops.
In the case where there are loops, one can explore a similar transmission of an interior
variable.
\bigskip
\begin{example}[Eigenvector centrality]
  A well known characteristic of undirected graphs, widely used in
  sociological analyses, is the eigenvector
  centrality\cite{graphpaper}.  Promises form a directed graph as long
  as each agent accepts its data from its neighbours. In any graph
  that aggregates data from its neighbours, some neighbours may
  accumulate more from their neighbours than others.  Consider an
  arbitrary graph of agents formed by promising to share their
  interior variable $\psi(A_i)$. Since agents have several neighbours,
  in general, each agent sums the values as before, leading to an
  effective updating equation \beq A_{ij}(\pi^{(\pm)}) \psi(A_i)
  \mapsto \lambda\;\psi(A_i), \eeq where the effective adjacency
  matrix of the graph is the matrix of promises: \beq A_{ij} =
  \pi^{(-)}(A_j,A_i)\pi^{(+)}(A_i,A_j).  \eeq This is an eigenvalue
  equation with a non-negative matrix, and the eigenstate
  corresponding to its maximum eigenvalue is also non-negative. It
  represents the eigenvector centrality distribution\cite{graphpaper}.
  The topology of the graph constrains dynamics, leading to a set of
  eigenstates over the population $A_i$. In a static graph, this leads
  to a distribution of `wealth' over the agents that are best
  connected. However, the matrix equation assumes that there is a
  dynamic equilibrium for constant $A_{ij}$, which may not be true for
  active agents. Stability thus depends on how quickly promises may be
  changing and whether other promises interfere with the values
  $\psi(A_i)$.
\end{example}
\bigskip
More examples of calculations based on elementary and homogeneous assessment functions
may be found in \cite{galam}.

The difference one must keep in mind as one
scales up to more sophisticated agents, with considerable interior degrees of
freedom, is that the ability to calculate something meaningful depends on knowing
the individual $\alpha_A(\pi)$ functions. Even in the case where there is a single
godlike observer $A_O$ watching over the agents with a single $\alpha_O(\pi)$,
that is not the assessment that drives the dynamics, so one cannot substitute it
for the actual one. This is a major obstacle to modelling in promise theory.
Finally, even being able to calculate values for interior variables, the final
outcome depends on what agents choose to select. Evolution and selection are
both at work in promise graphs.

When approaching promise theoretic agents quantitatively, the goal
is not merely to turn it into something like a classical physics problem,
but rather to take into account the special semantics of agency.
What does knowing a function like $\psi(A_i)$ actually tell us about the
function of a collection of agents?
What do the quantitative changes mean, in the context of the problem.
We must therefore turn to look into the characterization of class discrimination.

\subsection{Philosophical note}

In their efforts to explain phenomena, early natural philosophers
began by postulating mechanical {\em forces} and {\em potentials} to
drive downstream behaviour deterministically. In Promise Theory
parlance, a deterministic driving force would correspond to an {\em
  imposition}, i.e. an attempt to induce cooperation, without a prior
invitation. This is a deontic explanation, framing change as an
obligation or a command that must be obeyed\footnote{This Western
  scientific view probably has its origins in religious authority.}.
On an elementary level, one does not expect to need an invitation
to propose a change, yet---viewed in another way---this is what
charges and contact forces actually conceal.

For an autonomous agent, upstream determinism violates the basic
premise of local downstream self-determination\cite{promisebook}. In Promise Theory, a
downstream agent has no obligation to be affected by an upstream
signal (though it may still promise to do so). In other words, $F=ma$
does not refer to a non-ignorable force, but rather to a willing
recipient.

Modern accounts of change often prefer statistical formulations, owing
to the observed complexity of phenomena.  The attitude to causal
determinism also changed with the discovery of quantum mechanics, and
non-linear systems, where downstream causation could not be determined
with certainty. However, rather than abandon the notion of upstream
imposition altogether, quantum mechanics invokes a mystical aspect to
the theory rather than abandon the concept of force (source) and
response. In quantum field theory, one works around this with explicit
models of messenger particles or fields that signal force through
probabilistic absorption. In either case, one effectively sidesteps
the idea of upstream authority to force change on downstream agents,
and the behaviour of autonomous agents is similar to behaviours seen in
quantum mechanics for this reason.

The purpose of Promise Theory then is to represent the dynamics and semantic behaviours
of {\em agents} together, generalizing the specific models used in physics
or biology.  The purpose of comparing it to other theoretical
frameworks is to bridge a gulf of understanding to more familiar
tools, including probability.
In Promise Theory, the semantics of influence may still be found in familiar concepts,
now reframed as promisable intentions.
\begin{itemize}
\item For a downstream offer, $\pi^{(+)}$: propensity, impulse, potential, force, charge, influence.
\item For a downstream acceptance, $\pi^{(-)}$: affinity, permeability,
  responsiveness, plasticity, pliability,
  (vs resistance, inertia).
\end{itemize}
The distinction is that all responses ultimately come from within an agent.
Large scale forces and potentials may still represent a form of memory of accumulated
promise responses.

Why do we need another theoretical framework?  The simple answer is
that we have specific models, which fail to recognize each other's
similarities, while generic probability is incomplete with respect to
semantics and dynamics.  By choosing a bespoke language to unify
semantics and dynamics, we hope to restore some of the explanatory
details that probabilistic descriptions make ambiguous\footnote{The
  statistical methods of the natural sciences are attached to a
  pantheon of already-reserved semantics associated with standardized
  phenomena. This is a party to which artificial systems are often not
  invited: dynamically similar phenomena do not always mix as
  friends. Promise Theory establishes a tailored language without a
  specific territory.}.

\bigskip
\section{Probability}

Given an unpredictable phenomenon, many researchers will rush for the
toolbox of statistics and probability to model it. Superficially, this
sounds like an obvious choice, but it is not without its
pitfalls. Turning a phenomenon into a story about probability may
introduce unnecessary abstraction, and ultimately involve ambiguity
about causal information.

Probability theory calculates profiles of the relative likelihood of
alternative outcomes, by counting events from a process that has
either already happened or which may be in the process of
happening. Profiles are represented formally as sets.  Events arise as
measurements within the observational process. However, in advanced
uses to probability, events are abstracted to represent hypotheses,
contexts, and even theoretical models.

Whereas probability describes events, Promise Theory principally describes the
constraints on agents.
Thus a probability refers to a snapshot of a process, while promises
describe possible or `intended' trajectories.
One of the natural uses for promise theoretic probability, in a
learning sense, is to assess what an agent is doing and try to infer
what it may have promised, which is something like the question of
inferring `laws of nature' on the basis of repeated behaviour.

Possible events are counted by an observer over some number of
repeated process or sequence of $N$ trials.  Trials may be of any
size.  To characterize a process, we need to know enough about it to
know all of the possible alternative outcomes it may exhibit.  We then
begin by forming a basis of spanning sets, and count events with
similar characteristics. The dimensionless ratio $p_i$ of sample size
to total size is the measure of the importance or relative standing of
a specific occurrence that we refer to as probability.

If we are counting past events as an invariant frequency profile, then
$N$ is assumed to be large in order to achieve greater `objective'
certainty.  In a frequentist treatment, trials are assumed to be
performed `in parallel', i.e. independently of one another and under
equivalent conditions. Trials may be called space-like separated and
time is unimportant. The result has the semantics of a controlled expectation, whose
accuracy is based on the size of $N$.

\begin{figure}[ht]
\begin{center}
\includegraphics[width=5cm]{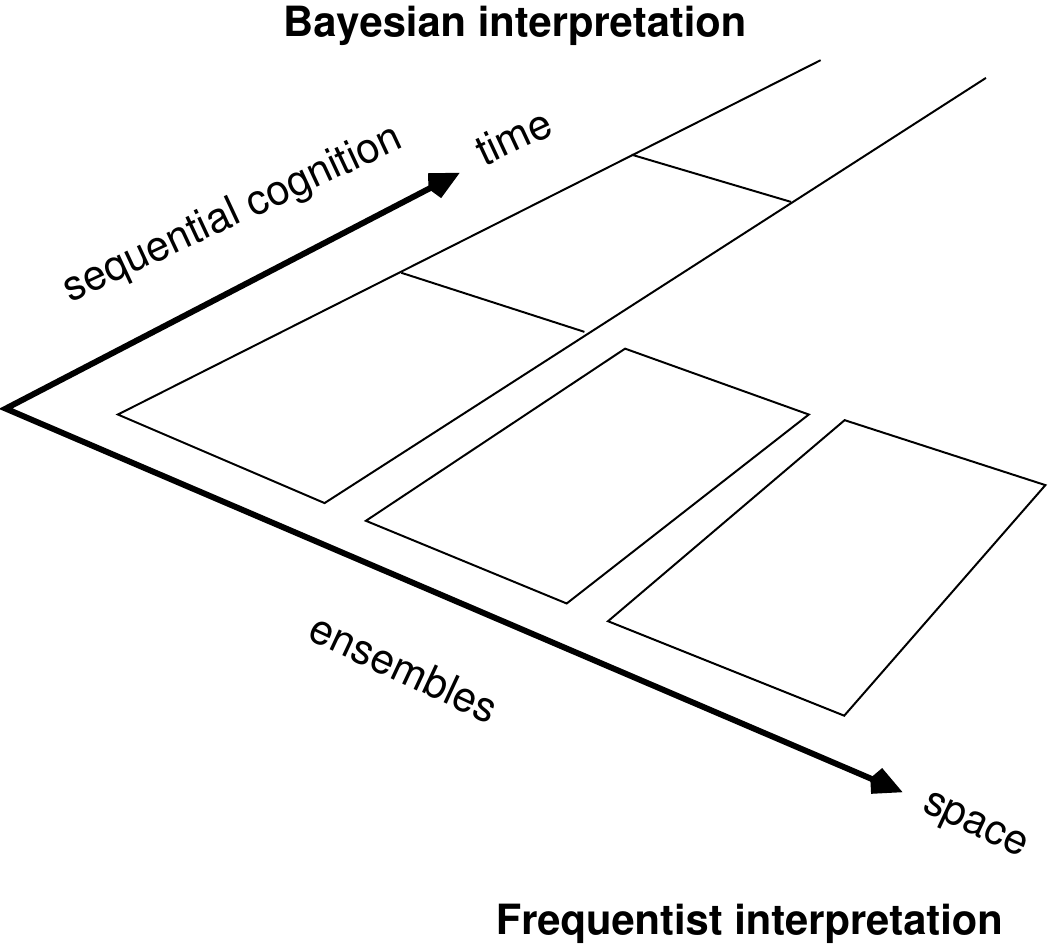}
\caption{\small Probability trials and ensembles. In a frequentist interpretation,
  trial ensembles are effectively space-like separated snapshots (transverse variation), whereas
  a Bayesian samples are ordered by a conditional precedence relation and thus
  represent time-like snapshots (longitudinal variation).\label{conchunks}}
\end{center}
\end{figure}

A Bayesian trial, by contrast, is taken in a time-like sequence, in
which the conditions of the experiment may be changing slowly, and we
are trying to update our estimate in some semi-controlled fashion (see
figure \ref{conchunks}). There is no built-in accuracy in this
approach. One speaks of changing `beliefs' rather than invariant frequencies. The
utility of the result depends on the integrity of the process one is
modelling rather than the method.

The logic of a Bayesian probabilistic reasoning quickly becomes maddeningly
complicated when trying to adjust fair but approximate estimates about
different outcomes likelihoods, on the arrival of new evidence (given that such
estimates are already uncertain) in order to minimize the risk of
making particular inferences. The following sections provide a brief review.

\begin{figure}[ht]
\begin{center}
\includegraphics[width=5cm]{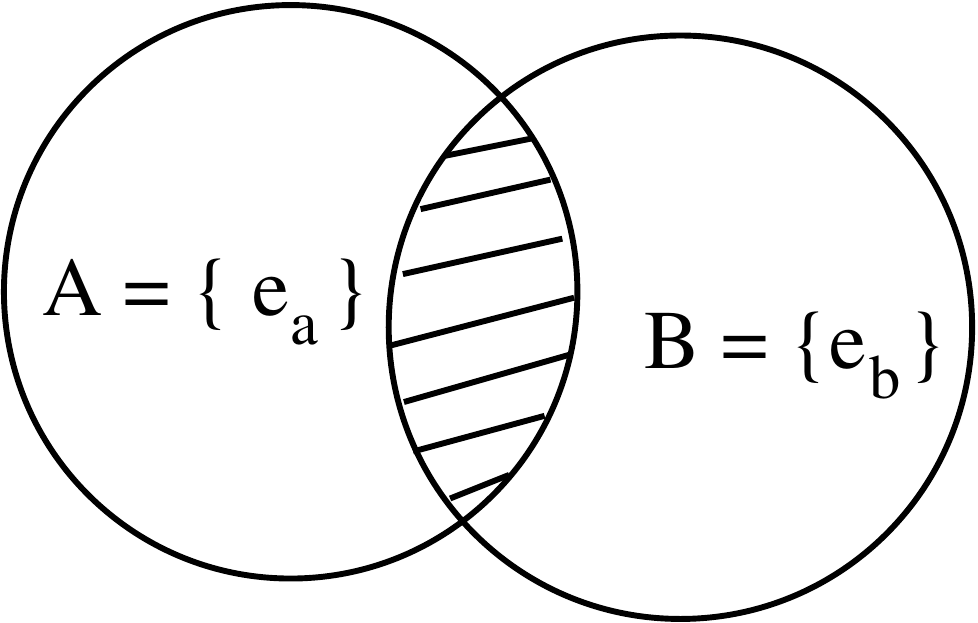}
\caption{\small Two sets formed from collections of events $A = \{e_a\}$ and $B=\{e_b\}$, and their
  overlap region $A \intersect B$.
  We note that the overlap region double counts coincident events.\label{overlap}}
\end{center}
\end{figure}

\subsection{Axioms}

Formally, any kind of probability reduces to studying the
function $P(e_a|e_b)$, where $e_a$ and $e_b$ are events, and the function represents the probability of
event $e_a$'s occurrence given the conditions implies by event $e_b$\cite{qm,pearl1}.
It is built on a few axioms and related results:
\beq
&1.& 0 \le\; P(e_a||e_b)\; \le 1\\
&2.& P(e_a||e_a) = 1\\
&3.& P(\neg e_a||e_b) = 1 - P(e_a||e_b)\\
&4.& P(e_a \AND e_b||e_c) = P(e_a||e_c)\,P(e_b||e_a \AND e_c)
\eeq
where $\neg e_a$ means NOT $e_a$, 
$e_a \AND e_b$ means $e_a$ AND $e_b$ (conjunction), and $e_a \OR e_b$ means $e_a$ OR $e_b$ (disjunction).
From these, we note that counting all elements in both sets must be corrected from double-counting the
overlap region:
\beq
P(e_a \OR e_b||e_c) = P(e_a||e_c) + P(e_b||e_c) - P(e_a\AND e_b||e_c).
\eeq
Moreover, if $P(e_a\AND e_b||C) = 0$, events $e_a$ and $e_b$ are
said to be independent, or mutually exclusive on the condition
$C$. For independent events
\beq
P(e_a\AND e_b||e_c) = P(e_a||e_c)\,P(e_b||e_c).
\eeq
Deciding whether or not sets overlap and by how much turns out to be
harder than one might think.  I depends on the definitions of the
sets (i.e. the invariants one has established), and having clear
definitions is sometimes a challenge that is not always met with due
care.

The subtle hurdle for probability lies in the problem of {\em normalization}.
How can non-local distributions over many agents be described by normalized
functions that rely on information they have no access to. Agents do not typically
know how many other agents are in the system: thus normalization is a transformation
performed by an aggregator and is not known to the system itself.

\subsection{Scale free and dimensionless}

Events form sets
$e_a \in \{ e_1,e_2,\ldots, e_D\}$ that may grow or shrink over time as sequences of events arrive
or are consumed by some process\cite{grimmett1}.
Each event adds one of $D$ different sets of
alternatives, where $D$ is assumed
constant. Given a number of arrivals, after a certain sampling time,
one forms a dimensionless ratio:
\beq
P(e_a) &=& \frac{N(e_a)}{N}\label{norm}\\
 N &=&\sum_a N(e_a).
 \eeq 
 By dividing by the sample size $N$, in the denominator, one
 `normalizes' the values relative to the sample size, and thus the
 probability is independent of its sample size. It is now a matter of
 `trust' that the ratio refers to a sufficiently large sample to form a
 meaningful distinction. We'll come back to this issue, since trust in
 probability is an important concern, and it has a specific meaning in
 Promise Theory.

In Bayesian thinking, one seldom writes simply $P(e_a)$, which implies an invariant property, since it tells
us little about the conditions under which the counting was done. Rather,
one may write:
\beq
P(e_a||C) = \frac{N(e_a||C)}{\sum_a N(e_a||C)},
\eeq
where $P(e_a||C)$ refers to a specific context or scenario $C$, rather than a
universally assumed phenomenon.
Form this we can test hypotheses by interpreting
this as a matter of inductive inference, in a Bayesian interpretation
of data collection. Thus $e_a$ and $C$ are now propositions and $P(e_a||C)$
is the degree of belief in $e_a$ given that $C$ is assumed true.
The price for this generalization is an implicit extension of the sample space to include a set of all
contexts, of which $C$ is a member. This can often be hand-waved away, but one must be
clear about the consequences of such handwaving.

In the context of a system of agents, where each agent learns a different
part of the observational picture, the challenge is to integrate these
different fragments into a meaningful whole. Normalization is one such challenge.
Since a dimensionless ratio retains no explicit information about the total
size of the sample, probabilities have the appearance of being scale
free. However, this is misleading, since the accuracy of the result
may still depend on the sample size in the denominator, and
convergence to an accurate value is also an assumption about the
asymptotic behaviour of the process being measured. Since there exist
stochastic processes that never converge, such as unconstrained
maximum entropy processes, a given probability may not be an accurate
guide to behaviour at all.  Knowledge of the normalizing denominator
is often the challenge to be overcome in calculating probability,
especially as the numbers change in response to new measurements.

An obvious quantitative question to ask is: how long does it take to establish
stable probabilistic significance? Identifying the appropriate timescale for
a phenomenon is a key starting point.

\subsection{Observables: event semantics}

The set of observables forms a spectrum of labels $a,b,\ldots$, which encode the
`semantics' expressed by observational events $e_a$. The
allowed observables form a set
$\Omega_E = \{ \omega_1, \omega_2, \ldots \omega_{D}\}$, and $D = \dim\Omega_E$, called the sample space.
This is assumed to be invariant. However, we can also consider multiple (assumed parallel)
processes with different sample spaces and combine them to form a single larger system.
This is
how probability becomes grounded with contextual conditional arguments.

Associated with each event is a model of perception that the observer
associates with the semantics of the event. The notation $\vec x_a$ is
associated with the vector of {\em features} associated with event $e_a$.
\begin{itemize}
\item In a physical experiment, dynamical variables may be associated with
a relatively small number of focused degrees of freedom, which can be modelled
on theoretical principles.

\item In modern machine learning models, feature vectors may have billions
of parameters.  The larger the number of features, the harder it
becomes distinguish a selection by probability alone, and thus
potentially to argue a causal connection.The amount of data needed to
discriminate alternatives is roughly proportional to the dimension of
this feature vector.  With insufficient data, one risks obtaining just
a few samples for edge cases. 
\end{itemize}
There is often a trade-off between semantic precision and available
data.  In the case where one believes that processes exhibit
longitudinal patterns, sample events can be combined into sequences
(sequences of DNA, sequences of words, etc), which are believed to
have repeatable significance, thus forming a `combined event'.  Short
combinatoric sequences of features may then be used to increase the
resolution of pattern recognition, at the expense of exponentially
fewer occurrences. The distribution of repeated sequences tends to form a power
law as a function of length\cite{burgess2020testingquantitativespacetimehypothesis1}.

\subsection{Event dynamics: the role of time}

Promise Theory can make use of
probabilistic concepts, and so we must seek to lay out that programme
in a coherent way.
Promises
refer to the space of semantics, or {\em intended behaviours}, which
include the classified behaviours expressed by a probability
sample space $\Omega_E$. They also go beyond these to express the context
of the process and extra-agent information such as why such
events are occurring.
Promises thus establish a criterion or a `guiderail' for
future outcomes.

\begin{figure}[ht]
\begin{center}
\includegraphics[width=6cm]{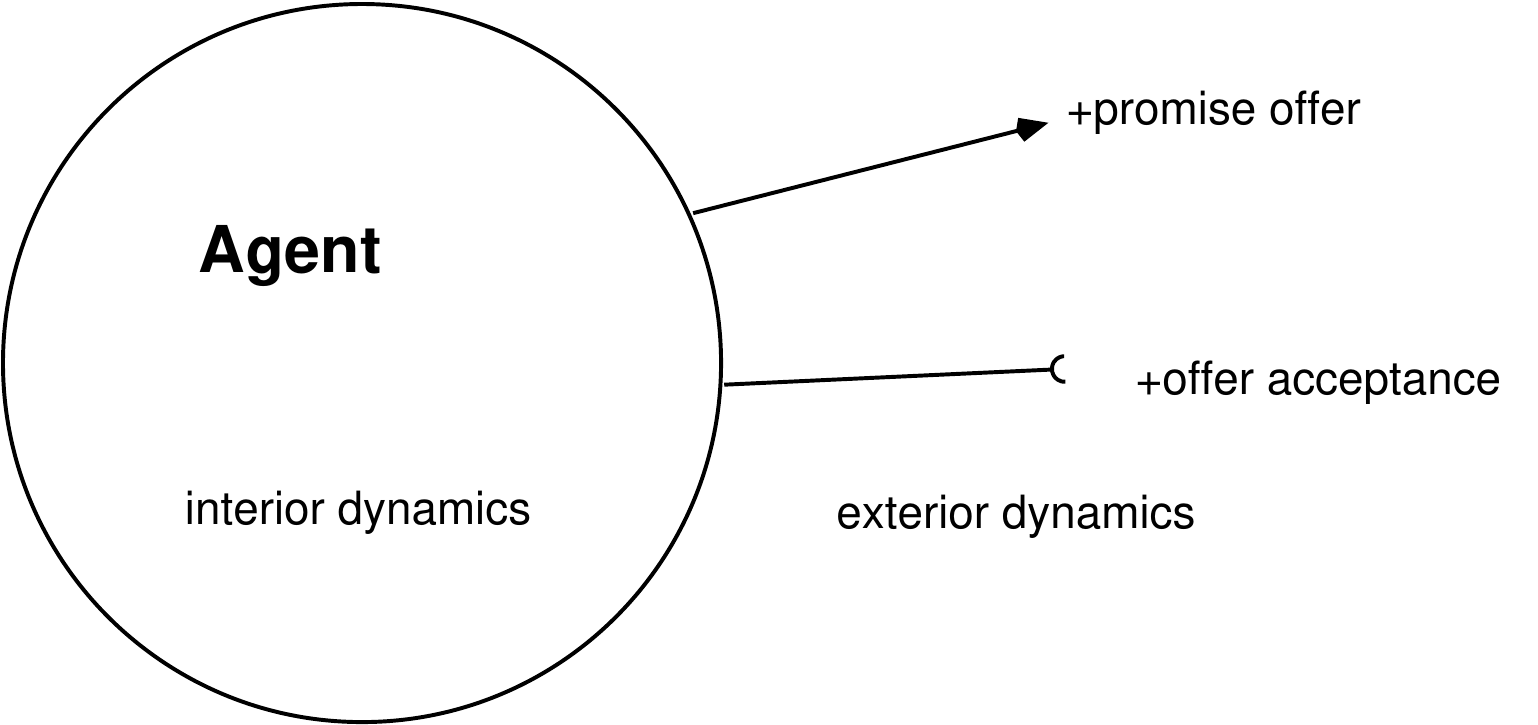}
\caption{\small Quantitative descriptions of promises may refer to
  interior dynamics, involved in keeping promises, or exterior dynamics
  involved in propagation of influence between agents.\label{agent}}
\end{center}
\end{figure}

If a dynamic variable cannot be predicted causally, one may still ask: can its
{\em probable} or {\em average} value at least be made approximately
causal?  This is where statistics and probability help to
give a formally precise algorithm for describing what may be inferred,
while actually remaining approximate.
The time snapshot picture of Bayesian probability suggests that we form
a strategy of separating variables into fast fluctuations and slow
(approximately constant or quasi-static) snapshots:
\beq
q(t) = \langle
q(t_\text{slow})\rangle + \delta q(t_\text{fast}),
\eeq
in which we characterize a slow
envelope of change by a running expectation value
$\langle q(t)\rangle = \frac{1}{\Delta t}\int_t^{t+\Delta t} dt \;p_aq_a(t)$, and a faster fluctuation part $\delta q(t)$ as
representing a `stochastic process' (meaning one we have given up
trying to explain and we now take on trust to be modelled by a
probability distribution).  Such slowly varying averages cannot be at
perfect equilibrium, but they may form persistent or approximately
steady states, so the tools of non-equilibrium effective theories are
in play, and energy conservation and other dynamical book keeping
variables may not be relied upon.

As we move from differential representation to a probabilistic
description, forces turn into outcomes, and rates of change turn into
the probability of a transitional change occurring.
\bigskip
\beq
\text{charge $\times$ potential} &\mapsto& \text{probability $\times$ effect}\\
\text{time derivative } \frac{\partial}{\partial t} &\mapsto& \text{probability $\times$ timescale}.
\eeq
This change of representation often happens without much fanfare or explanation.

\bigskip
\subsection{Discrimination by coincidence: jointly observed events}

The probability that $e_a$ happens together with $e_b$ within a given
sample is represented by the joint probability:
\beq
P(e_a,e_b) &=& P(e_a \intersect e_b)\\
&=& \frac{N(e_a \:\intersect\; e_b)}{N_\text{tot}}.
\eeq
$N(e_a\intersect e'_b)$ means the number of trials in which $e_a$ AND
$e'_b$ are observed together. One should remember that the $a$ and
$b$ trials do not have to run over the same set of outcomes, e.g. two sensors
might sample cars of the same type at different locations (same sample
space for each independent measurement), or two sensors in the same
location might measure type and colour (different sample spaces for
the independent measurements).
This is the same concept as the logical conjuction AND, but a glib treatment easily conceals the nature
of the process by which the counting is done, which is potentially confusing.

Temporal arrangement is important to the causal independence of events.
If, in a single measurement, the two samples are taken from different sources, i.e.
a sample has events of two independent processes occurring within
a single frame of measurement, where a frame may be space-like or
time-like separated, then $P(e_a,e_b) = P(e_a \intersect e_b) = P(e_a)P(e_b)$.
If however, the events were taken from the same source in the same time-frame, one after the
other then we must write
$P(e_a,e_b) = P(e_a \intersect e_b) = P(e_a||e_b)P(e_b)$\footnote{The question of whether the events $e_a$ and $e_b$ belong to the same
set is confusing. Regardless of the nature of the values representing the events
$e_a$ and $e_b$, simultaneous independent events form different sets (which may still overlap).
If, however, the events are from the same source, they are a single set.}.

The joint probability
$P(e_a,e'_b)$ is the probability that events $e_a$ and $e'_b$ occur
together, i.e. ``simultaneously'' for all intents and purposes with
respect to the trials. These measurements would refer to different
processes, i.e. measurements of different aspects of a system of many
possible aspects. For example, a joint probability for what the left
eye and right eyes see at the same time. For completeness of the observable sets,
it follows that one can always obtain the probability of each event alone by
summing over all possible outcomes for the other, resulting in the `marginal'
probability (a particularly unhelpful name):
\beq
P(e_a) &=& \sum_{b=1}^{D_b} P(e_a,e'_b)\label{margin}\\
\text{e.g.} &=&  P(e_a,e'_b) +  P(e_a,\neg e'_b)
\eeq
where $b = 1 \ldots D_b$, and $\neg e'_b$ is the complement of $e'_b$, or colloquially meaning `NOT $e'_b$'.
Problems and confusions abound when measures that are assumed to be invariant are
assumed to be representable by probabilities.

Taken together, joint observations have consistent sample space, which is the product space
of the sample spaces. This is consistent as long as every observer has the same sampling capabilities.
A colour blind observer would not register the same outcomes as one with perfect colour vision, for instance.
This tells us that we must carefully control what different agents interpret about each others'
data. Coincidence is also the phenomenon behind co-activation, in which an inference depends on
two things happening together. This is the basis for Hebbian learning\cite{hebb1}.

\subsection{Conditional events and reasoning in sequences}

If, within a single trial, the selection of a second event from a sample space
is affected by the selection of a preceding first event, then the probability of
selecting the latter event is said to be
conditional on the first. He result is an association between the two
\beq
P(\text{value}||\text{key}),
\eeq
like a `directory' or `lookup table' for key-value pairs, in which
the value is the likelihood for different associations.
In terms of semantics, we could visualize a process
of learning and updating such associative beliefs as a result of observation as a process of correcting
a telephone directory or a table of sines and logarithms, based on observing discrepancies.
The owner of the book updates the values by making notes in the original trained data\cite{bayes}.

As Pearl notes\cite{pearl1}, discussions of probabilities $P(A||B) = p$
have very different semantics than the more common logical notion of
reasoning $\text{IF } B \text{ THEN } A$, with likelihood $p$. The latter
favours an optimistic view: it is a policy to institute $A$ under some
threshold of evidence about $B$.  The former conditional probability
implies no policy; it only measures a degree of evidence for certain
inferences to be made about $A$ given an observation of $B$.  This
semantic gap is also the familiar deficiency levelled at semantic
knowledge graphs, which focus on assessed importance rather than
weight of evidence.

We abstract this into a story about random events as follows.
If two events are observed simultaneously, we can interpret this as one given that
the other is also present. If we know for certain that one of the events is
present, then we can reduce the overlap sample space so that the probability
is increased by removing the alternatives we know to be false. This is written
with vertical bars `$||$', thus the probability of $e_a$ given $e_b$:
\beq
P(e_a || e'_b) =  \frac{P(e_a,e'_b)}{P(e'_b)}=  \frac{P(e_a\intersect e'_b)}{P(e'_b)}\label{b1}.
\eeq
A conditional dependence suggests that
$e'_b$ must happen before $e_a$, which hints at causation. However, it  is not necessarily
enough to determine causation: that depends on what the events represent, and thus it
requires a semantic determination, not merely a potentially misleading notation.
Moreover, the assumed condition is symmetrical:
\beq
P(e_a,e'_b) = P(e_a || e'_b)P(e_b) =  P(e'_b || e_a)P(e_a).
\eeq
This is because we have already assumed that both are present in the overlap region
by virtue of $P(e_a,e_b)$ being non-zero. Thus, there is no implicit
order to these events in this formalism. They are simultaneous.
By combining this with (\ref{margin}), we sum to obtain
\beq
P(e_a) &=& \sum_{b=1}^{D'} P(e_a || e'_b)\,P(e'_b).\\
P(e'_b) &=& \sum_{a=1}^{D} P(e'_b || e_a)\,P(e_a)\label{b2}.
\eeq

From this symmetry, combining (\ref{b1}) with (\ref{b2}),
we have Bayes Theorem and the basis of Bayesian methodology:
\beq
P(e_a || e'_b) =  \frac{P(e'_b||e_a) P(e_a)}{\sum_{b=1}^{D} P(e'_b || e_b)\,P(e_b)}.
\eeq
Instead of interpreting this as about joint, simultaneous measurement events, Bayesian
methodologists often abstract the second variable $e'_b$ to represent a
different kind of set: a set that describes a
`context' for the measurement, i.e. a state referring to a new frame in
a process of discovery, as a state of knowledge in a model or hypothesis to be refined.
A context is still an event, but now it attains a higher significance as
a trigger event that explains the first event, by attaching to it with a some
probability. Our task is then to find which possible associated event
correctly identifies its partner event\footnote{Not that this could be a possible
cause, but could also just be a consequence of some other causal explanation.}. 
Relabelling (\ref{b1})
\beq
P(\text{posterior}) &=& \frac{P(e|H)P(H)}{P(e)}\\
 &=& \frac{P(x|\omega_i)P(\omega_i)}{P(x)}\\
&=&\frac{\text{Likelihood}}{\text{Evidence}} P(\text{Prior}).\label{Bayes}
\eeq
where the roles played by the terms are used, liberally,
to mean evidence $e$ for a hypothesis $H$, or
equivalently some feature attribute vector $\vec x$
associated with a classification of a `state of the world' $\omega_i$.
The terms are colloquially referred to as follows:
\begin{itemize}
\item Prior $P(H)$or $P(\omega_i)$:
  the probability of each hypothesized classification occurring, without qualification, from
  all prior experience. This is a distribution over states $\omega_i$ that we are trying to determine
  by measuring new scenarios.
\item Likelihood $P(e|H)$ or $P(\vec x|\omega_i)$:
  the probability distribution, all else being equal, that the classification $H$ or $\omega_i$
  is associated with the event $e$ or feature $x$. This is evidence about the properties of different events.
\item Evidence $P(e)$: the relative prevalence of a particular kind of event $e$ amongst all events, which
  is independent of the hypothesis and so plays a minor normalizing role.
\item Posterior $P(H|e)$: the generator of decisions, i.e. the probability that the given hypothesis
  or state of the world is $H$, given that we measured an event $e$.
\end{itemize}
The event $e$ (or its measured feature attribute $x$) is now a pointer to a
frame of evidence (quickly varying measurement), updated in sequence to refine
a model hypothesis $H$ (slowly varying and converging). 
$P(e|H)$ is viewed as the probability of an event $e$ occurring in the context of $H$.
If we start out with some prior estimate for $H$, as a
frequency distribution, that each new point refines. The refined estimate of $P(H)$ now maps
to a `causally' updated quantity $P(H|e)$, which is formally downstream of $e$.

One may thus define a probability
of some outcome relative to a model $P(e|M)$, in which the outcome $e$ ranges over a number
of possible values, and $M$ is an entirely different kind of object to $e$, perhaps one that cannot
even be enumerated. By writing $M$ like this, we denote the existence of a set of events (in which the model
is true and not true) as a dependency. In order to make sense, the model has to be an invariant.

\bigskip
\begin{example}[Inference of intent]
Could we, observing over time, infer the intention behind the behaviours of another agent, in the absence of
a promise? Or could one establish that the agent's promise is deliberately deceptive? Is the
agent lying? Lying needn't be a human trait. Animals that mimic one another often lie
in nature. Looking like a poisonous insect in order to ward off predators is a lie, with
respect to the acceptance promise that a predator is attuned to. A virus lies to
a cell receptor in order to reprogram the cell for its own intent.
\end{example}

\subsection{Sequential updating of estimates}

The Bayes formula is suitable for iterating over a sequence of incoming events
with features $x$. On the arrival
of the $n+1$th event, our belief in the model hypothesis would be updated to:
\beq
P^{n+1}(H) &\equiv& P(H||e_1,\ldots e_{n+1})\\
&=& \frac{P(e_{n+1}||H,e_1,\ldots,e_n)\,P(H|e1,\ldots e_n)}{P(e_{n+1}||e_1,\ldots,e_n)}\nonumber\\
\eeq
Thus one bends the meaning of
data to express a form of quasi-causal inference. This involves some complications
for continuous streams of data\cite{bayesianinference}.

The schema of the model is the set of pure states, $\omega_i$. It doesn't change
very often or very fast, assuming that the initial training data $D$,
which may be a set of topics per chapter in a book, or table of logarithms
for the schema that forms the basis set $i$, spanning intermediate events $w_i$.
\beq
P(x_0 | x,D) = \sum_i  \; P(x_0 || x,w_i) \, P(w_i || D)
\eeq
Each round effectively updates $D$ as well as updating $x_0$, through
the intermediate values $w_i$, for the schema of pure states labelled by $i$.

\begin{figure}[ht]
\begin{center}
\includegraphics[width=8cm]{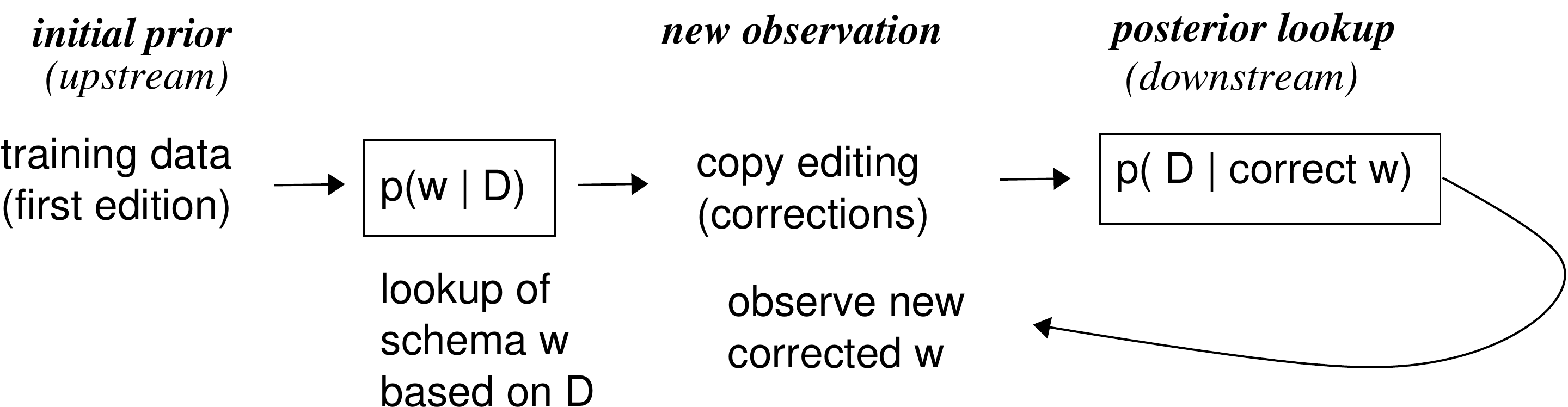}
\caption{\small Semantically, the corrective pipeline in a Bayesian learning process
  is a bit like copy editing a lookup-table or information directory.
  One starts with a ``first edition'' or training set of data, then
  one enters into a copy editing phase of updating the text observation by
  observation. The schema of attributes associated with each value(key)
  is approximately constant, but the values $w$ are random events. What is
  a key and what is a value seem to change places/roles as we proceed
  to correct items that are already in the directory.\label{editing}}
\end{center}
\end{figure}
Finally, eliminating a co-occurrence (coincidence of attributes)
involves summing out the intermediate states:
a conditional probability with two dependencies (co-activation)
becomes conditional on a single variable,
\beq
P(e||e_b) = \sum_c P(e||e_b,e_c)P(e_b||e_c),
\eeq
for example, where $e_c$ takes on the role of a context, chapter heading or
other classification variable.

\bigskip
\begin{example}[Controlled forgetting (again)]
  If we return to the section (\ref{avgvar}) using the controlled
  forgetting policy $\lambda_i$ to represent the controlled learning (with
  complementary forgetting),
\beq
\overline x(t+1) = \lambda x(t+1) + (1-\lambda) \overline x(t)\label{xxx}
\eeq
we can equivalently couch this in Bayesian terms by dividing the variable
$x$ up into a set of class ranges $\omega_i$, so that the probability
we learn over the course of tracing the development of an agent
over $N$ data samples is $p_i = N(\omega_i)/N$. This decomposition is an extra
step in the calculation. We can now write $x = \sum_i  x_i\omega_i$, like a vector
over the spectral decomposition, so that:
\beq
\overline x \equiv \sum_i\; p_i x_i.
\eeq  
Notice that the normalization $N$ is now formally changing with time. However, suppose
we start with some initial idea about the value of $p_i(t=0)$ and call this
our prior estimate $P(\overline x)$: then the Bayes rule says:
\beq
P(\overline x(t+1) || x_i) = \frac{P(x_i || \overline x(t))\,P(\overline x(t))}{P(x_i)}
\eeq
So the original equation (\ref{xxx}) is now transformed into a problem
of updating the expectation value
(divided into classes) by altering the probability rather than by updating the value directly.
This indirection is not useful in this case, but it illustrates how we can shift the dynamics
from one view to another\footnote{This is somewhat analogous to the moving of time dependence in the
  Schr\"odinger versus Heisenberg representations.}. Now we let
\beq
\overline x(t+1) &=& \sum_i p_i(t+1)\,x_i,\\
\overline x(t) &=& \sum_i p_i(t)\,x_i,\\
p_i(t) &=& P_i(\overline x).
\eeq
so that
\beq
\sum_i p_i(t) x_i = \lambda x_j + (1-\lambda)\,\sum_k p_k(t) x_k,
\eeq
and comparing coefficients of $x_i$:
\beq
p_i(t+1) &=& \lambda \delta_{ij} + (1-\lambda)\,p_i\\
&=& \left( \frac{\lambda}{p_i}\,\delta_{ij} + (1-\lambda) \right) \; p_i(t).
\eeq
and therefore $P(x||\overline x)/p(x_i) \mapsto \left( \frac{\lambda}{p_i}\,\delta_{ij} + (1-\lambda) \right)$.
Although we {\em can} formulate problems in a Bayesian manner, there is no advantage to doing so
unless we believe there are solid grounds for separating the dynamics of a probability distribution
from the variable value itself. The price for doing so is that i) the formulation is more complicated
than a simple convex updating, and ii) we are now using probability more like a dimensionless
potential function than as a normalized ratio linked to statistics.
\end{example}

\bigskip Interior probability might be an effective representation of an
underlying dynamical process, that allows us to invoke general arguments about
optimization, based on entropy for example, particularly when we can't
inspect variables directly. We hope to know as much as possible, and
therefore try to minimize uncertainty or invoke optimization schemes;
however, one should always keep in mind that knowing an accurate
probability for some change is not the same as having an explanation
for the phenomenon.

\bigskip
\subsection{Generalized stochastic systems}

If it indeed makes sense to separate the evolution of probabilities from
other dynamic variables, as it does in quantum mechanics for example, then
one may try to model the time evolution of a system directly as a probability:
\beq
\frac{dp(x,t)}{dt} = f(x,t).\label{master}
\eeq
An equation of this form expresses the evolution of a distribution $p(x) \mapsto p(x,t)$
as it is updated by some hidden underlying processes.

What generates a time dependent change in a probability, physically? The Bayesian
update rule in (\ref{Bayes}) gives some clues: there are three parts
that could figure in such a change: estimates about the nature of the
process, independent changes to the population itself, or to the
experiment, the normalization etc.  Making this inference is
ultimately similar to making inferences about agent's {\em intent}
from bulk statistical evidence, i.e. from causal bias.

We have to be aware that this model of time dependent probability
may be our own fiction, constructed by an enthusiastic abstraction, rather than a true
reflection of the interior degrees of freedom of the agents concerned.
On a practical level, one can study equations
for the time development of probability simply by making it a function of time,
as in (\ref{master}). So-called
master equations are
such examples include the Lindblad and Redfield master equations,
which are approximations to non equilibrium steady state
processes in quantum systems. The Langevin equation and Wiener processes are associated
with stochastic motion\cite{reif1,grimmett1,wiener1}.

Stochastic equations are, however, necessarily idealizations on a number of levels.
The evolution of a statistical distribution as a function
of time involves changes to populations on an average scale, which in turn
implies that what we are calling {\em time} is itself now a coarse grained
variable, which is not the fundamental unit of change\footnote{The same criticism
can be applied to quantum equations of motion, which involve several inconsistent
semantic assumptions about time.}.

\bigskip
\subsection{Semantics of autonomous pipelines}

We should not assume coherent behaviour from a collection autonomous
agents, without first setting up the conditions for cooperative order
throughout the collective. Bypassing autonomy exacts a price.

\begin{figure}[ht]
\begin{center}
\includegraphics[width=7cm]{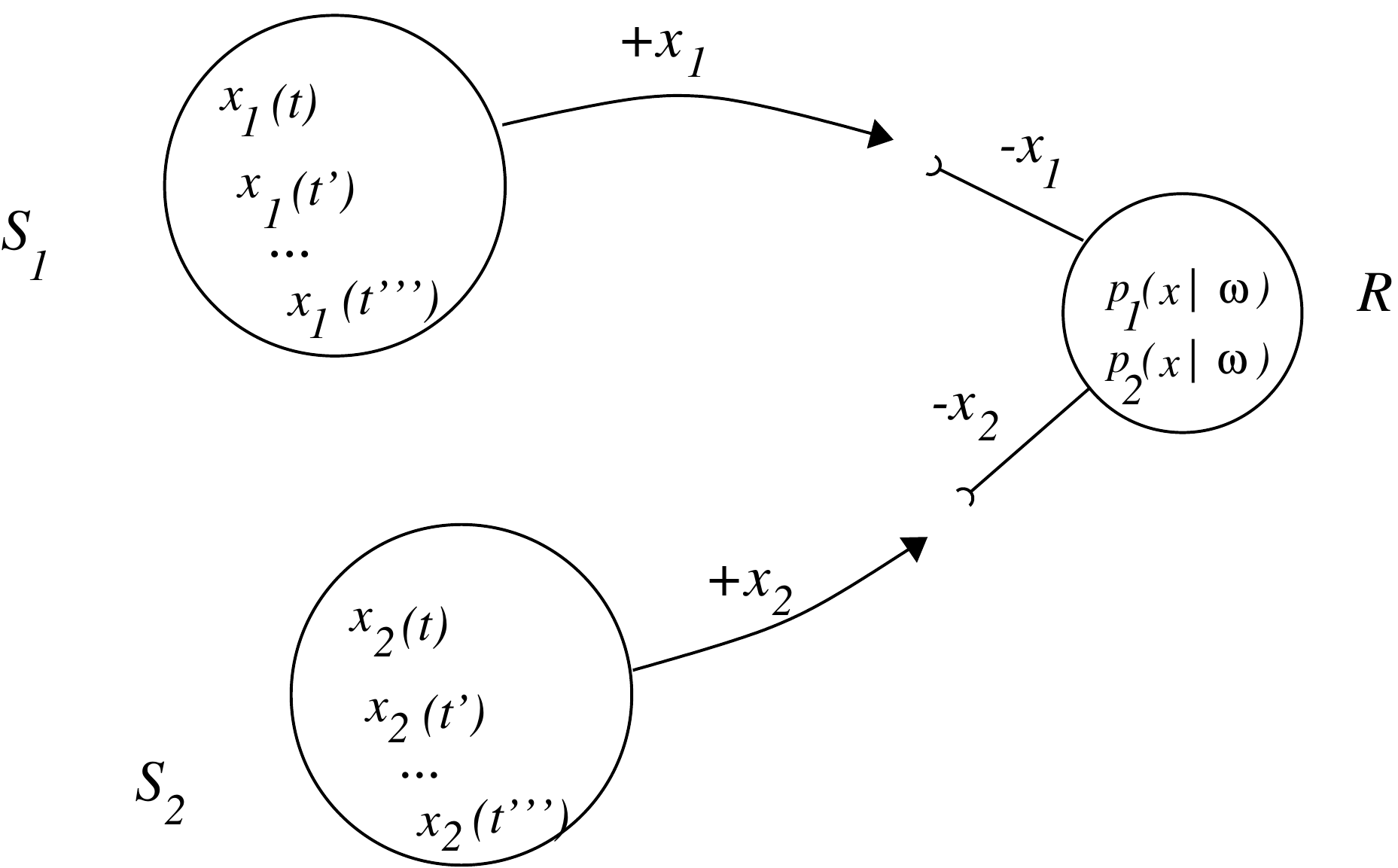}
\caption{\small Over a number of independent agents, there are potentially many
  independent parameters to capture behaviour. One must not assume coherence
  of sources without promises that enable coherence to form, and we should be aware
  of how values are normalized autonomously.\label{source}}
\end{center}
\end{figure}

Suppose data from autonomous agents are collected by sampling along a
pipeline from a source to a receiver.  In the absence of a promise
from some upstream agent, a downstream agent only has access to its
own interior variables: data collection needs to secure access to data
by a promise from the source, where data resides, to an agent that
will aggregate, calibrate, and compute results.
For the sake of probability, agents' promises can be treated as independent variables, as an ensemble, or as a number of trials,
etc.  Data arrive at effectively random intervals, on a voluntary basis,
as each agent operates on its own schedule (according to its own
clock) and with its own rate of change.
By the receiver, data do not
arrive marshalled into neat comparable batches, suitable for training, as
probability textbooks assume.  The downstream
principle applies to aggregators too (see figure \ref{source}).

Suppose we want a trusted data set $x_1, x_2, \ldots x_D$, to arrive as a set of events.
In Promise Theory, there are several prerequisites for this to happen: two promises
must be kept and an assessment must be completed.

Reliable agents interactions involve promise bindings of {\em offer} $(+)$ and {\em acceptance} $(-)$
to data pass from one agent to another. The source
of data must be identified: do the data values come from a single agent or from multiple
agents. Suppose each data value comes from a different agent $S_i$, $i = 1, 2,\ldots, N(S_i)$,
and that a single aggregator or receiver agent $R$ will collect data and calculate
statistical results.
Each source must promise access to its variables:
\beq
\pi_i(S_i) : S_i \promise{+x_i} R,
\eeq
similarly, $R$ must promise to accept all such values faithfully without truncating or
rejecting certain values:
\beq
\pi_R(R) : R \promise{-x_R} S_i
\eeq
where $x_R \intersect x_i = x_i$.
The receiver $R$ treats each data arrival as an event $e_a(S_i)=x_i$
and forms an assessment of the value $\alpha_R(x_a)$, which is the data value it perceives
from the source. It can now compute an average value from all its sources $S_i$:
\beq
\overline x = \frac{1}{N(S_i)} \sum_{a=1}^{N(S_i)}\;\alpha_R(x_a).
\eeq
We've assumed that $R$ only collects one such value from each $S_i$ over the course
of this calculation, so some synchronization is involved.

\subsection{Aggregation, confluence}

The foregoing steps were entirely in the domain of a single agent's promises.
Going further, we can allow a receiver $R$ to collect multiple data variables, from each source $S_i$,
and categorize the value into class buckets $\omega_a$,
which could also be different for each source agent: $\omega_a(S_i)$;
thus counting several events $e_a(S_i)$ from
each source in order to form a probability mass or density function for each agent separately
\beq
p_i(x_a|\omega_b(S_i)).
\eeq
The receiver can independently promise to calibrate the classes from each source leaving a simpler
result $p_i(x_a|\omega_b)$ which represents the variation in behaviours from a single
agent, and thus a single variable over multiple trials (over time). This assumes the addition of several
new promises, however.

The probability of events, in a Promise Theory framework, is thus really the probability of an agent forming
a certain assessment after establishing a process between them.
\beq
  P_{SR} \left( \alpha_R\left(\pi^{(-)}_R\, |\, \pi^{(+)}_S\right) \;||\; \pi^{(+)}_S,\pi^{(-)}_R \right)
\eeq
Although there is only a single probability data event, this is not a single event
  of agent actions, but a cluster.
In this sense, both promises acts like {\em event contexts} or {\em model hypotheses} in the Bayesian sense, which may or may not be
present before the events are measured by $R$.
Notice that this version of the conditional probability depends only
on $R$'s downstream assessment and its rate of change. Thus the agent
$R$ (downstream) is our sole lens for what is measured.

Finally, we should also bear in mind that, in a non-deterministic
world, with possibly unreliable agents, it is perfectly possible for events to
occur that have not been promised. In this case, they are a `surprise'
to the receiver. The receiver has the responsibility to marshal events
into a consistent timeline\cite{andras}.

What happens for a single agent interaction, in isolation, makes no reference to
whatever encumbrances it may have as a result of promising to other
agents. If an agent has to keep other promises at the same time, with
finite resources, this may affect its ability to keep that promise.
Thus, when we put an agent in a network of promises, this
might exclude certain options in its sample space $\Omega$.

Pursuing an analogy with quantum mechanics, it's natural to ask
whether we might calculate the pure states of a constrained agent
involved in some collective system, based on boundary conditions (see also section \ref{densitymatrix}).
Unlike quantum mechanics, this will not be a geometrical constraint
over spacetime that leads to a basis set of supportable
eigenfunctions. Where more complex semantics dominate, with a graph
topology, geometry takes a back seat and other process based criteria
take the lead in deciding what is allowed.  For example, in a
biological cell, what can be promised is based on biogenetic processes
which are the result of molecular bindings.

\subsection{Steady state behaviours}

When promises are static, i.e. independent of time over some epoch,
then agents may be said to be operating under steady state conditions.
Since a promise has the status of a constraint on the behaviour of the
promiser, i.e. a causal prerequisite
for events associated with keeping to some condition,
we want to write expressions like the following for an agent $A$:
\beq
P(\text{event } e_i(A) \;||\; \text{promise }\pi(A)),
\eeq
where the agent's usual events and its promises are different classes of happenings.

\begin{figure}[ht]
\begin{center}
\includegraphics[width=7cm]{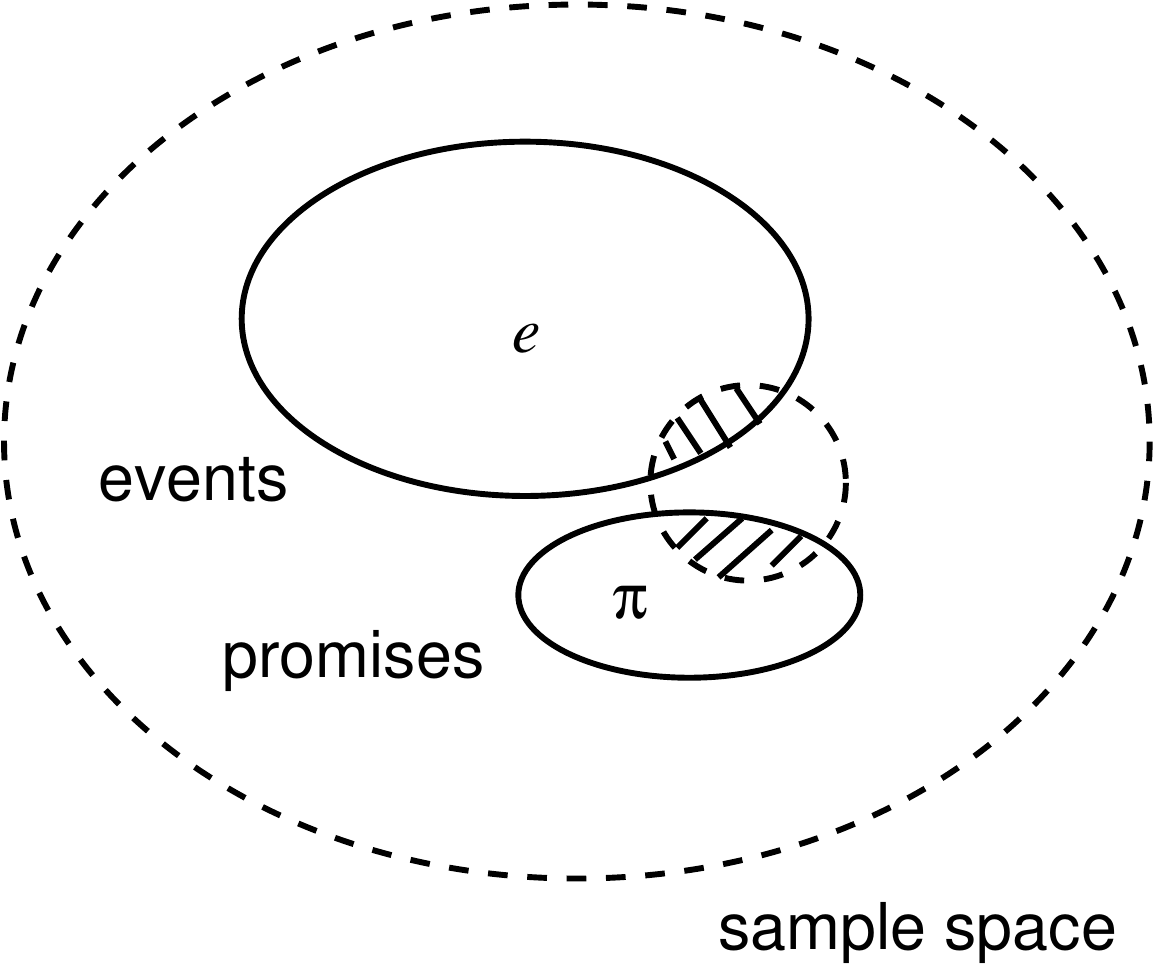}
\caption{\small In a promise based system, the sample space is formally enlarged to accommodate
  promise events that live alongside the pure states of the agent's dynamical activity.\label{associated}}
\end{center}
\end{figure}

From the foregoing discussion it should be clear that promises signal an active {\em
  intent} to act, whereas a conditional probability is an entirely
passive observational condition; it does not imply a causal order.
Probabilistically, then, a promise must be represented as a {\em
  complementary} set of mandatory preliminary events, before the
events an agent produces as output to keep the promise, where the
promise labels are new, independent basis elements.
We thus extend the sample space to account for the existence or non-existence
of constraining promises:
\beq
\Omega(A) \mapsto \{e_a\} \otimes \{ \pi_1^{(\pm)}, \neg\pi_1^{(\pm)}\} \otimes \ldots
\eeq
for all promises (see figure \ref{associated}).

A similar extension applies to both the sender/promiser $(+)$ and the receiver/promisee $(-)$ agents.
In this sense, promises complicate the classification of an agent's state. It is no longer
simply a stochastic system with random behaviour, but something more like a quantum system
with a basis set of specific outcomes that we hope to represent somehow. To a first approximation,
one could simply count the downstream assessment as a single event.

More accurately, for each agent independently, 
Bayes scaling formula (\ref{Bayes}) can now be written in the form:
\beq
P(\pi \;||\; e_a) = \frac{P(e_a \;||\; \pi)P(\pi)}{P(e_a)},\label{aaa}
\eeq
where $\pi$ is a promise, which is now used as a classifying conditional, and $e_a$ is an event generated
by the promiser and observed by any observing agent.
Reading between the lines, we understand these probabilities to be assessments $\alpha_R(e_a)$, $\alpha_R(\pi)$, etc,
made by some observer $R$, at the end of a data pipeline. The Bayesian contributions are then:
\begin{itemize}
\item The probability distribution of an observer's
assessment that the promiser is keeping its promise $P(e_a \;||\; \pi)$, given
observational data $e_a$, may be taken to be proportional to its prior
estimate that the promiser would do $e_a$ given that the promise $\pi$
was made.

\item The probability of the condition that the promise was given, $P(\pi)$, is
either 1 or 0 if the agent is known; however, if we are guessing
amongst a population of unfamiliar agents our estimate of how many
have made a promise may be any number between 0 and 1.

\item The probability
$P(e_a)$ that an agent might do $e_a$ unconditionally (irrespective of
whether it has promised or not), may also be guessed from data or
assumptions about a population. This is the usual Bayesian handwaving.
\end{itemize}
If we are looking at {\em propagation of influence}, along some path,
then at each transition we are really looking for the probability of a
promise binding resulting in a {\em shared event} as in (\ref{aaa}),
which includes a downstream assessment to accept this behaviour by
promisee $R$, given that it was offered by $S$ (see section \ref{freeenergy}). Moreover, the
probability of this unilateral assessment by $R$ is conditional on the
joint event of both the promises being present.

\subsection{Non-steady state behaviours}

Abrupt piecewise changes to a promise can be dealt with piece by
piece. When an agent's promises change slowly with time $\pi(t)$, we might not be
able to say much about the behaviour without a detailed interior
knowledge of agents, but we can look to non-equilibrium physics for inspiration.

Promise changes would normally be slow, so that a separation of variables can be performed to
identify a series of epochs over which promises are approximately
constant. The probabilistic snapshots, $P(t||\pi)$, will then lag
behind those changes, by hysteresis, owing time is takes to offer, accept, and assess,
as well as collect probabilistic assessments on the coarse
grained timescale of the averages.

The pure states of our system are, of course, the unique
(non-degenerate, inequivalent) configurations of the system,
satisfying the promise constraints.  The status of a promise change is
more like a change in boundary conditions than a change of state. If
a promise body contains variable parameters, the magnitude of a promised
service might adjust slowly, but
semantic changes to promises are rarely smooth differentiable
gradients.  While it is not impossible for agents to step up or step
down activities in a controlled manner, promises are more likely to
appear and disappear instantaneously as events.

As a calculated propensity, such a change could be implemented as a time
dependent probability. However if we try to interpret probability over growing data sets,
we are likely to run into trouble: the conditions are no longer invariant.  As data streams
grow in total size, results may converge either to certainty or simply
noise, depending on the sensitivity. The alternative approach is to make use of the
forgetting rate policy in section \ref{avgvar}, where possible. If a probability and its sample space are
not based on invariants, then its interpretation is thrown into doubt\cite{bayesianinference,burgessDSOM2002,burgessC14}.

Following \cite{bayesianinference}, one might approximate with a factorized
solution method over the independent agents. Then, approximation techniques like {\em master equations}
might also be applied, in the case where one could separate a interior Hamiltonian
evolution of the agent from a set of states associated with external
systems (see section \ref{densitymatrix}). In order to apply the kind of highly constrained evolution
implied by a master equation, one would have to identify all the pure
states of each agent, and then a set of separable jump operators.

The major challenge, in all scenarios, remains the lack of homogeneity between
agents. Agent chemistry, in practice, is often closer to molecular
chemistry than to a homogeneous thermodynamic material. Computer simulation
is the obvious choice for agent dynamics.

\bigskip
\subsection{Weighted pathways as intentions: trajectories and stories}

\begin{figure}[ht]
\begin{center}
\includegraphics[width=7.5cm]{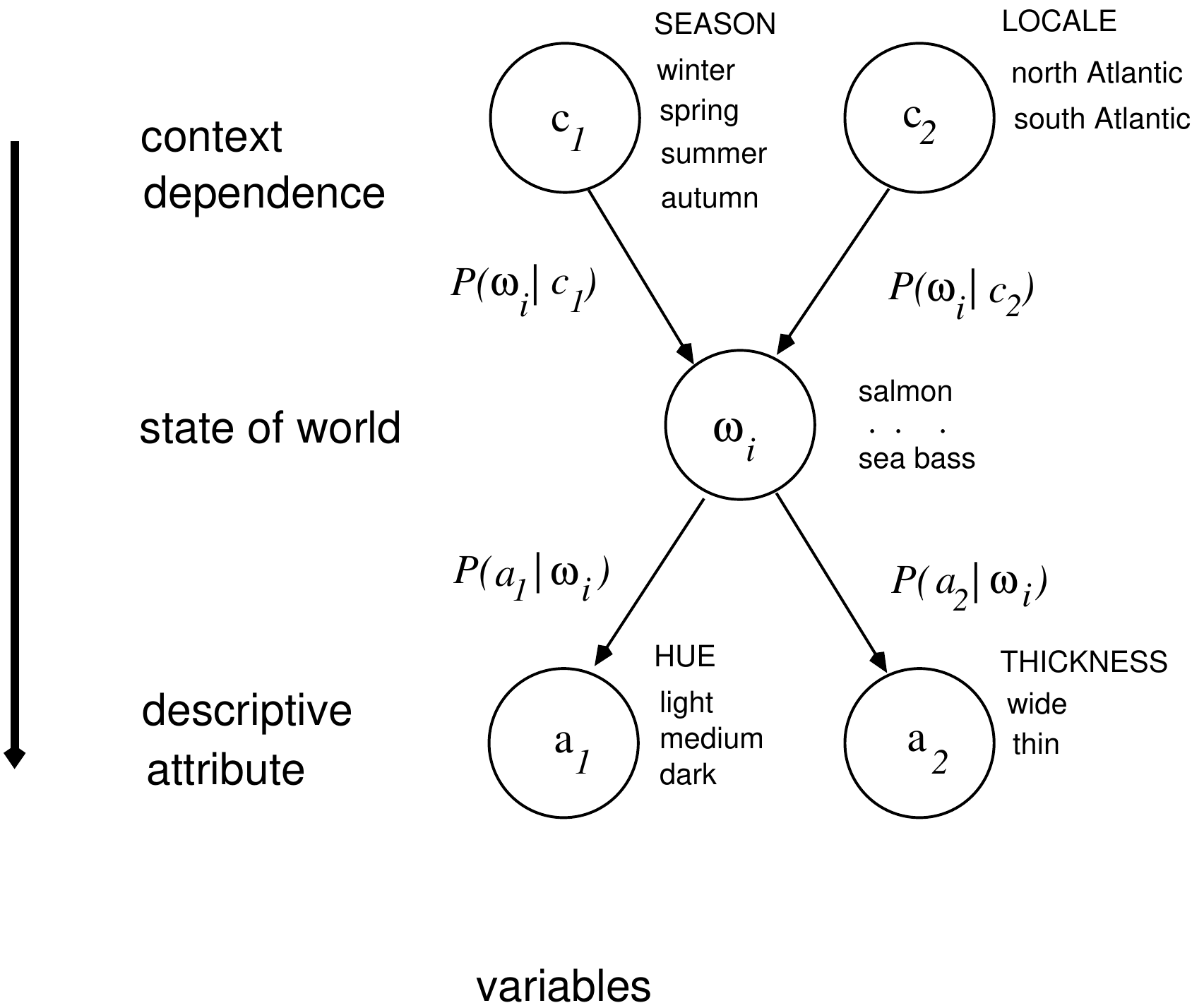}
\caption{\small Example of Bayesian network, from Duda et al. This also acts as a basic
  model for deconstructing semantics from a scenario: from context, to key states of nature,
  to attributes associated with them.
  The nodes of the graph are variables whose values
  play different roles in an inference.\cite{duda1}.\label{duda}}
\end{center}
\end{figure}

Modelling the state of the world as a set of independent states only
takes us so far.  Applying the theory of languages\cite{lewis1} to state labels, we can generate
many more states by forming strings from elementary states, either transversely or
longitudinally, with respect to time.  The number of effective state
labels can then be extended by combinatorics, and a semantic branching
process grows a single path into an exponentially large set of
distinct paths of a given length.
Although a chain of associative reasoning has no limit in length, in principle, the shortest
flexible version has three stages.
A simple model of triples is often used to represent contextualized
scene description\cite{duda1} (see figure \ref{duda}):
\begin{quote}
  ~\\
  (context,state,attribute)\\
  ~
\end{quote}
The way agents perceive a scenario goes along way to defining the semantics of a model.
Perception and semantics are naturally related:
\bigskip
\begin{quote}
A context or scene $\mapsto$ A set of characteristics (states) $\mapsto$ Associated attributes
\end{quote}
\bigskip
These downstream inferences correspond broadly to questions:
\bigskip
\begin{quote}
Where, When? $\mapsto$ What? $\mapsto$ How?
\end{quote}
\bigskip The effect of mapping this network is to create a Scene
Description Language\cite{scenedescription}, from whence more complex
language presumably emerges.  For example, if one thinks of arriving
on the scene of a murder mystery investigation, detectives first try
to capture a snapshot of the scene. Their job begins with describing
the evidence accurately. Later this associative chain is turned into
an network of inferences, by using past experience to assign
importances. Motives about {\em why} may be assigned effective
probabilities, as Bayesian beliefs, to complete the required semantics
of an explanation. Thus each semantic concept maps to a temporary assignment
of $P(x|\text{context})$

Probabilities may be assigned to the network on the basis of whatever process supplies
observational signal: e.g. from a mechanical finite state machine at the technological end of the
spectrum, to observational inference at the natural end.
Each distinguishable story, associated with an explanation of a scene, refers to selections from
a graph like the one in figure \ref{duda}\footnote{The explosion of pathways through the state space
is familiar from game theory\cite{rapoport1}. The normal form of a game refers to complete
coarse-grained strategies, while the extensive form of the game involves every possible
sequence of moves by all players\cite{myerson1}. Some of these might be equivalent
as a result of symmetries in the branchings. The same problem applies to discriminating intent by inference:
clearly any two strategies that lead to the same outcome have equivalent intent.}.

The paths through a graph of events represents a network of possible conditional
behaviours, representable as effective promises. Whether in terms of promises or
straightforward probabilities for events, the pathways develop the initial sample space
into a combinatoric sorting structure: leading us to
sequences that are not merely random but contain detailed semantic value. This
ability to further classify combinations of states, in a hierarchical
structure, is the motivation for taxonomy and ontology in knowledge
representations\cite{ontologies,fanizzi1,dietz}.

In practice, many events may fall into overlapping categories to a
significant degree---leading to a `classically forbidden'
superposition of pure states\footnote{I call such states, platypus states, since the platypus violates
the `natural ontology' by being an egg-laying mammal.}. For agents, there is no particular
problem with this: it will lead to the same need to make a selection,
formally. This is a choice.

The potential pitfall, when trying to construct a formal ontology for
a scene representation, lies in attempting to find a unique mutually
exclusive set of classes for a temporary phenomenon. The temporary nature of the
task suggests that a stigmergic approach to memorializing the information
is useful, because committing something temporary to long term internal memory
would be wasteful and counter-productive.

Computationally, the longer a sequence of events gets, the less likely
it is to be repeated or be equivalent to another sequence, so the
paths will represent unique phenomena.  The number of these, i.e. the
probable occurrence of a pattern, typically falls off like a negative
power law with
length\cite{burgess2020testingquantitativespacetimehypothesis1}. The
collection of shortest digital events are therefore not the answer to
understanding intentionality (which probably goes a long way to
explaining the evolution of language as a reasoning tool): finding
unique proper names for every possibility is an inefficient use of
memory resources. Hierarchical pattern models of meaning are more useful here
because they allow generalization and inference to approximate intent,
thus compressing the power law into a number of longitudinal steps.

The more intricate a semantic model for
classification, the more fragile it becomes to being broken by edge
cases. This is where the inexplicable abilities of human language to
represent meaning have outshone ontologies.

\bigskip
\subsection{Optimizing information with a least certainty principle} \label{information}

Maximum likelihood arguments and least certainty principles, are
associated with propensities derived from Information
Theory\cite{shannon1,cover1,bishop1}.  These are now commonly used to
talk about probability inference in artificial systems as well as
evolutionary biological structures.

Information theory characterizes probability
distributions and explores strategies for making minimal assumptions
(e.g. maximizing uncertainty subject to constraints). The Shannon
entropies\cite{shannon1,cover1} act as generating functionals for this
optimization. Such optima are not based on data inputs, but rather on
arguments about maximum likelihood---and thus, they are propensity
arguments.

The negative logarithm of base $D$, of
the probability of obtaining a symbol $i$ is called the information $I$ (sometimes called the
`surprise' or even `surprisal') in a message. It
measures the average number of alphabetic symbols
per unit length of message from an alphabet that classifies
event classes $\omega_i$ in a sample space.
\beq
I = -\log_D p_i.
\eeq
The expectation value of this information is the Shannon entropy:
\beq
S = - \sum_{i=1}^D \; p_i \log p_i,
\eeq
is an information density per unit length of message, and is a theoretical limit
on how much a message may be compressed by factoring and reducing symbols.

Although Shannon defined information as the average properties of
messages, it was based on the obvious notion that it is really the
digital {\em symbols} (or the semantic pure states), not their
averages, that constitutes the bread and butter of information.
Defining information purely in terms of probabilities is somewhat
misleading, and quickly leads one down paths the confuse knowledge
representation with inference\footnote{It's a bit like saying that an
  apple is only defined as a possible configuration of a fruit
  basket. One could take that point of view, but it misses the
  point.}.

Shannon's entropy measures are interesting because they act as generators for probability
distributions, and we can use them to compare the shapes of distributions in a number
of ways. So---to the extent that behavioural patterns are encoded in distributions of
actual behaviours---one can use information theoretic forms of entropy to measure the
relative shapes of such patterns. However, this method has no checks and balances built in,
and the scale-free dimensionless nature of probability allows one to compare
highly inappropriate quantities as well as appropriate ones. It can be used as a kind of
inference methodology, as long as one keeps control of the data manually.
The basic Shannon entropy of a distribution is an average measure of its bias over some distribution of pure
states. Minimizing the intentional bias, or maximizing the uncertainty corresponds to a principle
for hedging probabilistic strategies.

Consider a probability distribution $p_i$, where $i = 1,\ldots, D$.
A completely unbiased distribution is flat, i.e. $p_i$ is constant
and equal for all $i$. This is called the unconstrained maximum entropy distribution.
The opposite case of a completely biased distribution in which there
is a single $p_i = 1$, and all other $p_{j\not=i}=0$  (the Dirac delta distribution) is the
unconstrained minimum entropy distribution.

Entropy thus acts as a generator for constraining the probability mass function.
One can maximize it, subject to different constraints, to find the distribution with the
least amount of bias. This is independent of the meaning of the probability. It's actually
a property of the dimensionless ratios.

If one thinks of a probability distribution (with a fixed number of categories $D$) 
as a characterizing the statistical composition of a message formed from a sequence of events.
The message is thus viewed as consisting of an alphabet of $D$ symbols.
The Shannon entropy then acts as a generating function for distributions.
The unconstrained maximum entropy distribution satisfying
\beq
\frac{\delta S}{\delta p} = 0,
\eeq
leads to the maximum entropy distribution $p_i = \text{const}$.
One may add certain additional constraints using the method
of Lagrange multipliers. 
One can add to this constraints such as the normalization condition:
$\sum_i p_i = 1$:
\beq
S_\text{norm} = \alpha \; \left( \sum_{i=1}^D \; p_i - 1\right)
\eeq
This is already hinting that normalization is a possibly problematic constraint.
The constraint of constant expectation value for some quantity $E_i$ with
the Boltzmann distribution:
\beq
S_\text{Boltz} = \beta \; \left( \sum_{i=1}^D \; (p_i E_i - \langle E)\rangle\right).
\eeq
The solution to this problem is sometimes called a Boltzmann mixed
state\footnote{Note that, in this usage, state refers to an ensemble macrostate,
not a microstate of an ensemble component.} in physics,
since it mixes different $E_i$ observable outcomes into an ensemble
average.
\beq
p_i = \frac{e^{-\beta E_i}}{\sum_j e^{-\beta E_j}},
\eeq
and for thermal systems, once identifies the inverse temperature $\beta=1/kT$.
The Gaussian distribution is also a maximum entropy distribution, for probability
with a constant proximity to a single average value\cite{treatise1}.

\bigskip
\begin{example}[Swarm behaviour]
  In earlier work I speculated that a swarm is a collection of autonomous agents
  that acts to minimize its informational entropy\cite{siriAIMS1}, i.e. that
  the cohesion of the swarm (over any promised attributes) is characterized by
  a tendency for agents to align with one another. In this sense, it is a generalization
  of the old $XY$ model of spins, with a generalized interpretation of spin and some
  random selection. In swarms, 80 percent of a swarm may try to stay together, while
  the remainder go off at random to explore new possibilities. The latter possibility is what
  discovers new dynamical configurations and prevents the swarm from crystallizing into a
  ferromagnetic state.
  
  There are two key variables for the agents in a swarm: their direction of flight and their
  inter-agent separation. For example, in two dimensions, we could describe classes
  of direction by an angle class $\theta_i(t)/2\pi$, where $\theta_i(t)$ is the velocity direction of
  agent $A_i$. If we assume that the agents fly with approximately constant
  speed, then the velocity distribution is like $v_i=vp_i$
  and the probability $p_i^{(\theta)}$ of flying in this direction has entropy $S_\theta$.
  Similarly, the average distance between agents $p(x_i)$ is distributed around some fixed distance
  that maximizes the safety of the agents: too close may risk collision, while too far risks
  losing cohesion for navigation or attack from predators. The maximum entropy distribution
  for distance subject to $\int p(x)(dx)=1$ and $\int (dx) (x-\mu)^2 p(x)=0$, for each $x_i$
  is
  \beq
  p(x_i,t) &\mapsto& {\cal N}(x_i,\mu(t),\sigma^2(t))\\
  {\cal N}(x,\mu,\sigma^2) &\equiv& \frac{1}{\sqrt{2\pi\sigma^2}}e^{-\frac{1}{2}(x(t)-\mu)^2/\sigma^2}.\label{gauss}
  \eeq
  where ${\cal N}(x,\mu\sigma^2)$ is the Gaussian (normal) distribution, which we define here for future
  reference.
\end{example}
Thus, we have turned a dynamical problem into a problem about
probabilities, though not based on data---based on a variational
argument.  This may be regarded as propensity for the inter-agent
distance in a three dimensional space, at an instance in time $t$,
where $\mu(t)$ is an average position of its nearest neighbours in the
whole swarm. In order to calculate this individually for each agent
$A_i$, every agent must promise to be visible and each $A_i$ must be
able to sense the position of its neighbours.  As a probability
calculation, this is beyond the capabilities of simple agents, but as
a predictive principle based on avoidance, it is a plausible
prediction.

\bigskip

\bigskip
\subsection{Convergence of intent: Predictive Coding and Active Inference}\label{freeenergy}

The model of `Predictive Coding', in neuroscience, has attracted a lot of attention in
connection with inference and pattern recognition in neural
systems\cite{predictivecoding1,predictivecoding2}. It is a
probabilistic approach to perception.  A variant of this
called Active
Inference\cite{friston1,friston2,friston4,friston3,freeenergytutorial},
proposes a variational approach based on so-called `Free Energy
Principle', which can be interpreted in a number of different ways.
Pragmatically, it is an approximation, which may be used to solve for
a fixed point by using a variational parameter.  In that sense, it is
version of Feynman's variational method for statistical
mechanics\cite{feynmansm}.  More subtly, by applying Bayes theorem to
invert the order of dependency in a sensory data pipeline, it turns
inference into an argument about the {\em convergence} of sensory experience,
somewhat like the CFEngine concept of convergence\cite{burgessC11}.

The philosophy behind the approach is the suggestion that organisms
seek out a local fixed point equilibrium of information about the world.
Imagine that an agent, or set of agents, observes the world and builds a model
of the world in terms of associations between sensory stimuli and internal
interpretations. The hypothesis goes that organism's cognitive faculties
essentially promise a convergence of this key-value association 
based on a minimization of Shannon's joint entropy. The idea is similar to that proposed
in \cite{siriAIMS1} for swarm behaviour.  A critique of the method may be found in
\cite{cao2020newlabels} and an extensive thesis investigation in
\cite{millidgethesis}. An example, which reveals what is actually
going on between the beautified arguments, may be found in
\cite{freeenergytutorial}.
which can be approached either by
altering the world or by altering an average model.  

We suppose the purpose of a brain is to converge to a model which
minimizes the predictive `error' for interpreting what is
senses. Since error is subjective and defined somewhat vaguely, it
ends up being a relative moving target. However, as a single
mechanism, it might account for many phenomena.  In short, the
suggestion is that perception is not purely the bottom-up aggregation
of sensory promises $(+)$, but actually a top-down selection by
acceptance promises $(-)$ of a prior representation of the concept,
which drives prediction. Recognition can only be based on a dictionary
of existing concepts.

The hypothesis has been transformed into a discussion of
Bayesian probability distributions, which makes it familiar to a wide
range of researchers, but which does not try to explain how it might
be realized by agents, such as neurons. I'll not attempt to resolve
this issue here, but it seems plausible that Promise Theory's
autonomous agents might help to do so. We can begin by reformulating
the arguments in terms of Promise Theory concepts.

\bigskip
\begin{example}[The Predictive Coding approach]
According to the method of active inference, the goal of a learning `brain' is to minimize
the semantic distance between external observation and interior model
and thus infer a classification.
where the semantic distance is measured by the relative entropy,
expressed in the form of the Kullback-Liebler divergence.
\beq
D_{KL}(Q|P) = \int dQ \; Q\log \left(\frac{Q}{P}\right).
\eeq
The proposal is broadly that, over time, predictions and empirical behaviours converge so as to
minimize the relative entropy of the behavioural probability $Q^*(x||M)$.
The brain is supposed to have a generative model of a scenario, expressed in probabilities $P(x||M)$.
The variables $x$ are called `actions' here---they would correspond to `intentions' in a promise
theoretic model. They correspond to the behavioural variables of the model.
The goal of predictive coding, and active inference, is to fit an estimate of the posterior distribution $Q(x||M)$ to the actual
posterior $P(x||M)$, by functional minimization of:
\beq
 \min_{\lambda} D_{KL}(Q(x||M,\lambda)\;||\;P(x||M)),
\eeq
The convergence of the solution may be learned iteratively, layer by layer, in a neural processor and may be adjusted
on either distribution, e.g. by learning new behaviours $P$ or by changing behaviours $Q$.
This observation is perhaps the core of the idea.
Expressed in terms of Bayesian probability, the idea has a number of problems.
When $D_{KL} = 0$, the variational distribution equals the `true' distribution
and one has solved the inference problem---though without a clear definition of
a correct answer. It solves the position of the swarm even while the swarm is
still gyrating.

Since the ideal minimization is potentially difficult, one attempts a variational approximation.
The variational `free energy' $\cal F$\footnote{The thermodynamical roots of the free energy concept refer to a
characterization of the amount of activity in a system that is
available for doing work\cite{reif1,zemansky1}. In the case of
perception, work means the process of arriving at an
inference.  Entropy is that part of the total energy which has been distributed
into useless background noise. It only adds to the uncertainty of the signal
remaining `free energy'. In Friston's free energy conception, probabilities do not have
the dimensions of thermodynamic entropy or energy, but they do capture the same spirit: that of
a focused `accounting variable', which is what energy does in a
different context.} reverses the Bayesian order to obtain an upper bound
on the answer\cite{predictivecoding2}.
\beq
&\,&D_{KL}(Q(x||M,\lambda)\;||\;P(x||M))\nonumber\\
&=& D_{KL}(Q(x||M,\lambda)\;||\;\frac{P(M,x)}{P(M)})\nonumber\\
&=& D_{KL}(Q(x||M,\lambda)\;||\;{P(M,x)}) + \log P(M)\nonumber\\
&\le& D_{KL}(Q(x||M,\lambda)\;||\;{P(M,x)})\nonumber\\
&\equiv& {\cal F}(\lambda).\label{upper}
\eeq
where $P$ is the generative model of probability for actions $x$, $Q$ is the variational fit,
and $M$ is the presumptive model. As an agent problem,
learning may be thought of as a kind of swarm of virtual agent neurons, which
are flying with some synchronicity. Each individual agent's probability could be modelled
as a maximum entropy Gaussian, like (\ref{gauss}) around some fluid configuration.
The expectation maximization algorithm then takes turns in optimizing
$\cal F$ with respect to $\lambda$, keeping $M$ fixed, then the other way around,
until there is best possible convergence.
Using probability as a key-value representation of a behaviour is a way of representing
its `shape' in a linear sample space.
As a computational model, this has been related to approaches to reinforcement learning\cite{bayesianinference,predictivecoding2}.
It's unclear whether that method is a fair representation of a brain.
\end{example}
\bigskip
The probabilistic arguments make the concept accessible to a wide audience, but conceal some pitfalls, when applied to agent
systems (as noted by \cite{freeenergytutorial}). A valid critique concerns the ability to compute distributions of
probability over a population, which is not managed from above---and more specifically, to calculate
the normalization of probabilities based on plausible signals (see equation (\ref{norm})), which is essential to all
Bayesian methods. As a dimensionless ratio, probability is a fully
scale-free characterization of relative importance, but belief cannot
be a scale free character as it would never be able to distinguish between scales.

Without a central controller to maintain shared information,
normalization across multiple agents is both problematic to calibrate and
calculate. Taking a probabilistic method at face value, each agent involved in a calculation
needs access to the total sample size.  A single stage aggregating
node, with $N$ inputs can (at least in principle) count the number of
active acceptance or receptor promises is has and use this as a local
count. However, if an agent is to calculate Bayesian rescalings,
imported from a multitude of agents, we have to be careful in assuming that
upstream values propagated by one agent are calibrated with the same
meaning as downstream agents (figure \ref{source}).

The same caveat applies when making top-down variational arguments
about the distribution of properties across a population of agents, as
in section \ref{information}, for instance to maximize or minimize
some uncertainty. Maximum likelihood distributions, which postulate a
cooperative organization across several agents have no causal
potency. They apply to each agent individually.  When invoking an
argument based on information, we therefore have to ask: what
behaviour in agents might lead to an effective tendency to optimize
behaviour effectively as a swarm?
A godlike observer, looking down on a flock of birds has no causal
influence over their behaviour, so its ability to calculate
inter-agent probabilities post hoc cannot be used as an explanation of
their behaviour.  The signalling between the agents is the causal
prerequisite. In this case, we may be looking for a potential argument
rather than a probability.

Let's apply the variational principle in a different way, following
a promise theoretic approach, which follows that laid out by \cite{freeenergytutorial}, but with a revised
narrative and a forgetting algorithm to replace Newton's
method. This serves as a neat summary of all the foregoing methodologies.
\bigskip
\begin{example}[Convergence approach]
  Suppose we have a sensor agent $S$, and a number of downstream
  agents that help with the processing of its input. We can try the
  active inference approach, this time using the CFEngine style
  minimal computation convergent method\cite{burgessC14}.  This aligns
  well with the method discussed in \cite{freeenergytutorial}.  We
  consider a small collection of specialized agents, each contributing
  to a calculation of an optimum key-value lookup.  The goal of this
  collaboration is to match sensory signals to approximate measures of
  size or distance. For example, in the example of
  \cite{freeenergytutorial}, this could be an animal looking at
  different items and trying to gauge the type of food, or birds
  flying close together in a swarm monitoring each other's positions,
  by guessing some feature $\phi$ such as the size or distance based
  on observed light.  A possibly non-linear transmutation of the
  signal, with dimensions going from from lumens to metres, may be
  involved in order to calibrate meaning.  There are many similar problems one can imagine in which
  one uses a learned key-value table to transform a sensory (prior)
  signal into a semantic (posterior) inference (see figure
  \ref{senses}).

  We formulate the problem of associating the correct inference of feature $\phi$ to the
  input signal $\tau$ by maximizing a likelihood function. It's not clear
  that this is implementable with elementary agents yet, but I'll come back to that. We
  could treat this maximum likelihood function either as a dimensional potential or as a
  dimensionless probability. Since
  everything can be kept linear, the details are not important, but
  there is a small semantic bonus to thinking about the problem in a
  Bayesian way. I'll follow the discussion of \cite{freeenergytutorial}\footnote{The notation has been altered in relation to \cite{freeenergytutorial}:
\beq
v \mapsto x  &-& \text{variable}\\
v_p \mapsto \langle x\rangle &-&\text{variable mean}\\
\Sigma_p \mapsto \sigma^2(x) &-&\text{variable variance}\\
g(v)  \mapsto T(x) &-&\text{variable to $\langle \tau\rangle$ converter}\\
u \mapsto \tau &-&\text{sensory sample potential}\\
\Sigma_\tau \mapsto \sigma^2(\tau) &-& \text{sample variance}
\eeq}.

The organism's brain contains a collection of autonomous agents associated with this model
of recognition, as well as a much larger part that we won't pretend to model. This small portion of agent circuity
will learn about the relationship between light and size or distance, as in \cite{freeenergytutorial}, by forming
an associative lookup capability, so that the module can promise to
decode a sensory potential and transform it into a measurement. Such an association could be modelled by any kind of joint
potential or joint probability function $p(\text{light},\text{size})$ or $V(\text{light},\text{size})$.
the discussion in \cite{freeenergytutorial} is instructive, as it highlights some of the difficulties
with measurement that elementary agents face. The arrangement of agents is shown, approximately in
figure \ref{senses}.

Light stimuli come from `without', i.e. outside the organism.
The function of the light sensor agent is to emit a signal $\tau$, given some input light:
\beq
\pi_S: S &\promise{-\text{light}}& \text{Environment}\\
       S &\promise{+ \tau \,|\, \text{light}}& L_i.
\eeq
where $L_i$ are the agents that deal with remembering light values. Without filling in all the details,
these agents must accept the signal and calculate the parts of an expression to predict
a direction for gradient ascent:
\beq
L_i &\promise{- \tau} S,\\
L_i &\promise{+\sigma^2(\tau) | \tau}& A'.
\eeq
making their own promises to calculate mean and variance (see equation (\ref{runav}) and the subsequent discussion).

Similarly, a set of agents $M_i$, dealing with the semantics of distance maintain an
interior model of how distance fits together with the light measurements.
\beq
M_i \promise{+\phi}
\eeq
So,
independently, from within the organism, there is a model of dynamical variables $x$, that will converge
to stable values $\phi$, representing features of the world, one of which is modelled as an example
here. The hypothesis now is that the function of the brain is to promise to associate the best value for this
feature with a sensory potential $\tau$. Clearly, this leaves many questions unanswered, but it's a beginning
to modelling the perception of the collective.

\begin{figure}[ht]
\begin{center}
\includegraphics[width=8cm]{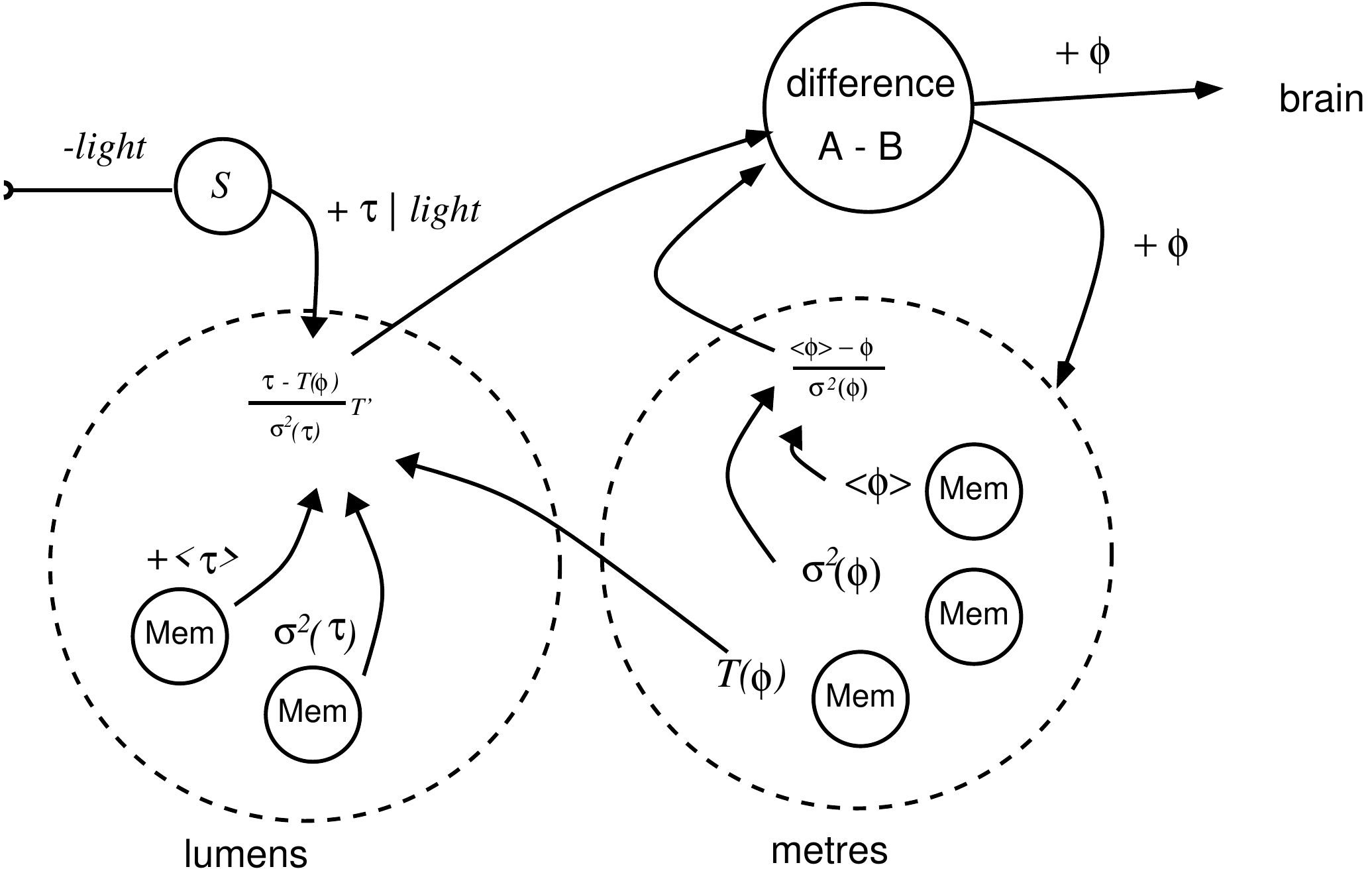}
\caption{\small The agents and promises in a Predictive Coding sensory inference pipeline, sketched roughly.
  Each circle is an agent. The stippled super agents clusters act as modules for partial
  memory centres where running averages are kept. An approximate `maximum likelihood' 
  estimate of the dynamical variable's feature $\phi$ is computed within a few
  sampling iterations. We assume that each agent is working to sample inputs and recompute
  outputs according to its own resources.\label{senses}}
\end{center}
\end{figure}

What the collective wants to know is: which $x$ yields the maximum likelihood of associating an input light signal
with internal representation $\tau$. We denote the best effort value of the feature $x\mapsto\phi$, and we'll
calculate it by maximizing an association, subject to some promise constraints:
\beq
\max p(\tau,\phi) \Huge|_\pi
\eeq
This represents a kind of search operation on the space of {\sc express}ed promised values\cite{burgess_sst2025},
subject to constraints that the final value $\phi$ converges on a single (hopefully correct) answer.
As autonomous agents, we imagine that every agent is doing its job at all times, at its own rate---as opposed
to a pipeline of data driven from the exterior light impulse. There is no single calculation resulting in the
devastating truth, but rather a continuously adjusting picture of the world as it happens.

Searching for the optimum involves several unknowns. If we use Bayes theorem, we can rewrite the search
as an asymmetric conditional probability, which leads to a method for using the free energy approximation to the result.
We are looking for an agent $A$ whose role is to promise to
take $\tau$ and return a $\phi$:
\beq
\pi_A : A \promise{ + \phi | \tau} A'\label{cc}
\eeq
This association can't be a Markov process,
like a simple function, as it will converge by depending on a memory of what it has seen in the past, and we expect
new information to update its memory of what is most likely too.
It makes sense to associate the conditional promise (\ref{cc}), i.e. $\pi_A(x | \tau)$ with a conditional
probability $P(x||\tau)$ (which we may also refer to as the {\em response function} $R(x||\tau)$ of the organism),
for a `feature vector' $x$.
If we solve for this instead, it effectively locates an upper bound
for the response function, as noted earlier in (\ref{upper}).  
The organism will already have some idea of what $x$ it may be looking at, and further
samples $\tau$ will correct that. I use the symbol $\tau$ for the events of the sensory input
signal, to indicate its role as driving the time-base of the pipeline.

While a Bayesian representation helps us to make sense of this
challenge, it's not clear that it would be implementable by a collective
of autonomous agents. A variational approach, called the Free Energy
Principle that uses Gaussian trial functions for the likely shape of
the response, offers a suitable method for autonomous agents, because
it is a maximum uncertainty function, centered around a mean
value. This also aligns with a minimum certainty principle as a worst
case scenario.

If we maximize an association response, based on a speculative trial
version of $x \mapsto \phi$, then we can construct a plausible
approximation to the response function at the maximum likelihood
solution.  The good news is that---despite Bayesian appearances---we don't need the whole distribution,
only the region of maximum likelihood, so we won't have to waste a lot
of memory remembering useless detail.  So our response function is,
applying Bayes' theorem: \beq R(x||\tau) = P(x||\tau) =
\frac{P(\tau||x)P(x)}{P(\tau)}, \eeq where $P(\tau||x)$ may be
interpreted the associative signal threshold for recognizing the
attribute $x$.  The `free energy' $F$ is defined as the negative
logarithm of the maximum likelihood response: \beq
-F(x,\tau) = L(x,\tau) &=& \log(P(x||\tau))\nonumber\\
&=& \log(x) + \log(P(\tau||x)) -\log(P(\tau)).\nonumber\\
\label{xxy}
\eeq
We want to maximize this with respect to $x\mapsto\phi$, and thus compute:
\beq
\frac{\partial L(\phi,\tau)}{\partial\phi} \mapsto 0.
\eeq
The last term in (\ref{xxy}) is independent of $\phi$, so we can ignore it.
The independent agents will behave simply, and with bounded uncertainty.
Using trial Gaussian functions for the first two terms of (\ref{xxy}):
\beq
P(x) &\mapsto& {\cal N}(x,\langle x\rangle,\sigma^2(x)) ~~~ \text{(memory)}\\
P(\tau||x) &\mapsto& {\cal N}(\tau,\langle T(x)\rangle,\sigma^2(\tau)) ~~~ \text{(sensory)}
\eeq
we can calibrate or map one normal response to another, like a key-value table.
where $x$ and $\tau$ have different measures or dimensions: $\tau$ is measured, say, in lumens,
and $x$ in metres.
$T(x)$ is a function or process that transmutes an agent's memory value of $x$ into a value
that can be compare to a sensory potential $\tau$ (i.e. with the same dimensions),
and we have defined $P(\tau||\phi)$ so as to seek a maximum response
where the converted `remembered value' of $T(\phi)$ matches the current input value $\tau$.
Substituting for ${\cal N}(x,\mu,\sigma^2)$ from equation (\ref{gauss}), and differentiating
\beq
\left. \frac{\partial L}{\partial x} = \left(\frac{\langle x\rangle -  x}{\sigma^2( x)}
+ \frac{\tau - T( x)}{\sigma^2(\tau)}\cdot\frac{\partial T( x)}{\partial  x}\right)\right|_{x=\phi}. 
\eeq
We want this to approach zero from our best estimate of $\phi$, and adjust it according to the
sensory potential $\tau$.
By writing this Bayes decomposition the `wrong way around', we are saying that our trial
value of $x$ will act as an attractor for the optimal answer, so that---after a few
iterations---the input $\tau$ and $ x$ will be a matching pair.

This is a complex expression for an elementary agent to compute exactly,
so it is implausible that a neuron could find an exact answer. However, as \cite{freeenergytutorial}
points out, this isn't strictly necessary. All we really need is the sign of updates to $\phi$
in response to $\tau$, and $\partial T/\partial x$ can probably be assumed to be strictly
positive. In addition, the dimensionless variables used by Friston in his original paper present a problem for
an elementary agent, because a standard deviation $\sigma$ requires a square root to be calculated,
where as $\sigma^2$ is merely additive. This may, in fact, be an advantage. We don't necessarily want a scale-free
representation as it would not be able to distinguish actual sizes.

The conclusion is that a simple agent only needs to emit two kinds of signal for
positive and negative updates, based on differences and overall multiplying scales. The rate of
convergence is unlikely to be optimal, but it should end up at a stable answer for $\phi$, which is yet to be calibrated
downstream of the cognitive process anyway.

So, if we imagine a time variable associated with samples of $\tau_i$, for $i=1,2,\ldots$ etc, then
given a prior estimate of $\phi(t_i)$ from memory, a new estimate can be checked
from observing with the sensor to produce $u(t_{i+1})$, then approximately following
the gradient towards the maximum:
\beq
\phi(\tau_{i+1}) = \phi(\tau_{i}) + \frac{\partial L}{\partial\phi}\;\Delta\phi^2,
\eeq
where $\Delta\phi^2$ is a small increment with the dimensions of $[\phi^2]$ to make the second
term small compared to the first, 
\beq
 \frac{\partial L}{\partial\phi}\;\Delta\phi^2 \simeq \text{sign}(\frac{\partial L}{\partial\phi}) \Delta\phi,
\eeq
and
\beq
\langle \phi(\tau_{i+1})\rangle = \lambda \phi(\tau_{i+1}) + (1-\lambda)\, \langle \phi(\tau_{i})\rangle,
\eeq
for some $\lambda$. If the increments
are small enough, this scale may be chosen as an arbitrary constant, since we only care about the sign
of the derivative.
\end{example}
This final remark underlines a basic lesson of Promise Theory studies: in order to
work in a classically predictive way, agents must make accurate promises, i.e. be able to
effectively represent quantities that are real valued numbers, with quite good accuracy.
For this to be true, the density of interior states must be sufficiently high to capture the phenomena
concerned.

\bigskip
\subsection{Trust as a running cognitive attention potential}

Trust is closely related to the semantics of promises for a reasoning agent. It is a mnemonic for
a process that summarizes how agents select promises from others and attend to their progress.
For completeness, I summarize the main findings of Promise Theory\cite{trustnotes,burgessdunbar2}.

Agents with interior memory may choose to monitor their interactions with one another. This
leads to an understanding of the basis for Dunbar's social brain
hypothesis\cite{dunbar1,dunbar2,burgessdunbar2}.  The assessment
functions an agent associates with promises from other agents are a
measure of the reliability in keeping promises---which is a measure of
the agent's trustworthiness. The assessed `probability' $V$ of keeping a
promise is a measure of an agent's trustworthiness with respect to
that promise is a form of invested work accounting, and can be turned into
a specific rate of kinetic work by straightforward dimensional
arguments---identical to those for energy conservation.

The rate of kinetic velocity of mistrust $v$ is determined
by dimensional analysis $\2mv^ 2 \propto V$. Thus, trust in Promise
Theory is a process of watching over other agents on a continuous
basis. It results in slowly varying potentials and
responses\cite{burgessdunbar2}.

Trustworthiness $V$ (process potential energy) is a summary of
historically accumulated work of alignment with past promises,
expressed as a coarse snapshot of the slowly-varying history. It has
the semantics of reliability or believability for each individual
agent to assess. Conversely, attentiveness (process kinetic energy) is
an immediate release of work at some rate or velocity, in response to
residual uncertainty, expressed directionally by the potential alignment. 

\def\Nav{\langle N\rangle_{\overline T}}

\begin{figure}[ht]
\begin{center}
\includegraphics[width=8cm]{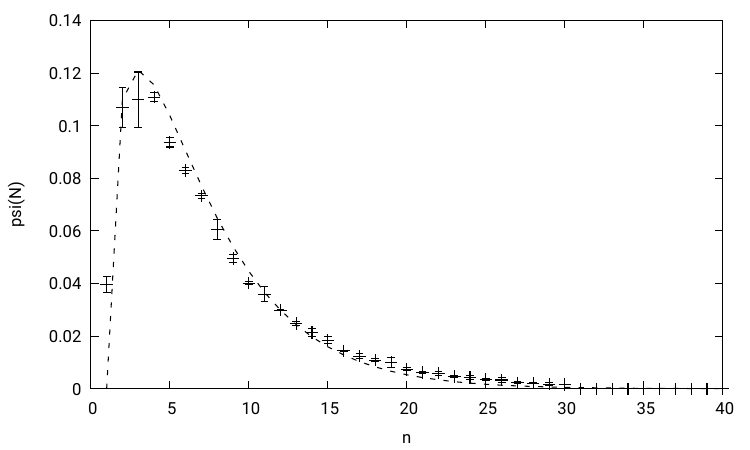}
\caption{\small Curve fit of data using the formula in equation \ref{formula} with a data
from around 200,000 agents on Wikipedia from \cite{burgessdunbar2}.
\label{datafit}}
\end{center}
\end{figure}

\bigskip
\begin{example}[Agent clustering and social group sizes]
The work of a single agent, interacting in a group of size $N$, would be expected to scale as
\beq
W\text{(agent)} = \frac{c_1+c_2(N-1)+\ldots}{c_0 N_\beta}
\eeq
where $c_0$, $c_1$, and $c_2$ are constants\cite{burgessdunbar2}.
At low utilization, we can expect the availability or channel capacity to be approximately proportional to
the number of agents interacting. Once contentions sets in, this effective number slows down
as agents begin to leave a group to an average---which is the value at which contention is maximal.
When $\beta E = 0$, the probability has to be 1, so for $N=1$ (self), all the share is
in one agent's hands. So $c_1=0$. Now we have a single scale $C\equiv c_2/c_0$ representing
the level of shared of contention between agents. To determine this, we use the promise
seed configuration again below. Note that, at maximum entropy, a group is evenly distributed
by definition, 
without favour to any particular agent, so based on these dimensional arguments, 
we expect the large $N$ limit to take the form of a Boltzmann distribution:
\beq
P(\beta) \sim \exp\left(-\frac{\chi(N-1)}{N_\beta}\right),
\eeq
where $N_\beta$ is some scale that characterizes the intra-group contention. Small $\chi$ 
implies tolerance of contention, or loose coupling and thus larger group sizes,
while large $\chi$ implies some kind of territorial overlap that leads to altercation.

We assume, based on maximum entropy arguments, that at large $N$, averaged over large ensembles
the probable work fraction $P(W)$ for distribution takes the form of a Boltzmann distribution over the relative
costs\cite{reif1,treatise1}:
\beq
P &\sim& e^{-\beta E}\\
\text{where (dimensionless)} ~~~ \beta E &\mapsto& \frac{W(\text{agent})}{\text{Total capacity}},
\eeq
where the availability is the finite budget for shared resource channel capacity. 
We recall, from Shannon\cite{shannon1}, that the channel capacity is a dimensionless representation 
of the power:
\beq
C = B \log \left( 1 + \frac{W(\text{agent})}{\text{Cost of contention}}\right)
\eeq
where $B$ is the maximum bandwidth for throughput, which is consistent with our assumption. 
With these points in mind, and assuming that interactions between group members are not `all at once', 
but interleaved approximately one at a time,
the accumulated work should be proportional to the group remainder size:
\beq
W_n \le \frac{W_\text{max}}{N}.
\eeq
We can take the cognitive capacity as a share for work:
\beq
(N-1)W_N = \2m v^2,
\eeq
for some rate $v$. Now, we arrange to measure these quantities in units such that we can
compare dimensionless ratios. In dimensionless form, we can compare the only matching scales
in the problem:
\beq
(N-1)\frac{W_N}{W_\text{max}} = \2 \frac{m}{m_\text{min}} \left(\frac{v}{v_\text{max}}\right)^2,
\eeq
We make the identification,
\beq
\frac{W_N}{W_\text{max}} \frac{m_\text{min}}{m} \equiv \frac{\beta}{\Nav}, \label{xx}
\eeq
which has the form
\beq
\frac{\text{Fractional work effort}}{\text{Fractional cost of involvement}}\times \text{efficiency},
\eeq
where we use the constant $\beta \le 1$ as an efficiency.
This is motivated by the identification of $\Nav$ as the scale for group size with
maximal contention cost.
From (\ref{xx}) we interpret the Dunbar group size as being based on:
\beq
\Nav = \text{fraction of work budget} \times \text{cognitive efficiency}.
\eeq
From data \cite{burgessdunbar1}, 
$\Nav$ is associated with data called the group contention cost, which is an emergent
scaling limit determined by looking to the group size at which contention arises, or when
$\Nav$ agents are all watching closely.
We can now predict the
dimensionless propensity for finding a group of size
$N$, by combining growth and decay, according to the equilibrium of the attachment and detachment rates. 
The probability balances probabilistic
attachment and detachment, shaped by a trust potential which is learned
by a past history memory process. Growth is initially by invested kinetic effort in aligned order $\sqrt{\nu}$
and decay is by average disordered contention $\exp({-\nu})$ (see figure \ref{datafit}).

\beq
\psi(\nu) &=& \frac{4}{\sqrt\pi}\; \frac{\nu^{\2}\; e^{-\nu}} {\langle N\rangle_{\overline T}},
\eeq
where
\beq
\nu &=& \frac{2\beta(N-1)} {\langle N\rangle_{\overline T}} ~,~ (N > 1),\label{formula}
\eeq
and $\beta$ corresponds to a dimensionless (probabilistic) rate of promise keeping.
\end{example}
\bigskip

It's quite rare to be in a position to compare theory with data in such a scenario, but here
we are fortunate \cite{burgessdunbar2}.
Here, we have a model of the most probable size of a social group
in a seeded collaboration of agents with human characteristics.

Trust obviously has some semantics associated with moral judgements, which may
use far more simplistic arguments, based on assessments made by some custodians
of truth, such as police or a court of law. It is trivial to make the following
Bayesian moral criteria:

\bigskip
\begin{example}[Honest and dishonest agents]
In general, the existence of a promise would be expected to increase the likelihood
of an event keeping the promise:
\beq
P(e_{\text{keep}(\pi)}||\pi(A_\text{honest})) \ge P(e_{\text{keep}(\pi)}||\neg\pi(A_\text{honest}))
\eeq
However a dishonest agent may behave in the opposite way.
\beq
P(e_{\text{keep}(\pi)}||\pi(A_\text{dishonest})) \le P(e_{\text{keep}(\pi)}||\neg\pi(A_\text{dishonest}))
\eeq
The trustworthiness is a generalization of this assessment.
\end{example}
\bigskip

\bigskip
\subsection{Semantic spacetime and paths}

The physical graph of promises between agents, as the spacetime of
active sources in a dynamic system, is mirrored by another graph: a
virtual realm of all the data variables $x_a(A_i)$, which live inside
the agents and may, in principle, be shared between them. These
variables and representations form a space called Semantic
Spacetime\cite{spacetime1,spacetime2,spacetime3}.
Like a probability distribution, it is unclear whether one could actually
construct this space from real data; but, as a speculative model of
observed properties, Semantic Spacetime is a representation of concepts
and ideas, which the agents may be considered to be `thinking
about'.

Semantic spacetime forms a directed graph, in which there are
four kinds of arrow\cite{burgess_attention,burgess_sst2025}: temporal
or causal arrows, spatial containment, descriptive attributes, and
semantic similarities. The arrows have explanatory semantics and may
also have weights, indicating their relative importance to downstream
processes.

The semantic spacetime construction is similar to (but goes beyond) Wright's decomposition rule\cite{bayes}
for causative correlations, using Reichenbach's common cause principle. For events, causation may imply
process graphs in which processes look somewhat like Feynman path integrals:
\begin{itemize}
\item $e_a$ causes $e_b$ directly,
\item $e_a$ causes $e_a$ directly,
\item $e_a$ and $e_b$ share a common cause.
\end{itemize}
Causation generally requires some confirmation by counterfactual inference:
\beq
P(e_a \intersect e_b) &>& P(e_a)P(e_b)\\
P(e_a \intersect e_b || e_c) &=& P(e_a||e_c)P(e_b||e_c)\\
P(e_a || e_c) &>& P(e_a || \neg e_c)\\
P(e_b || e_c) &>& P(e_b || \neg e_c).
\eeq
To apply Wright's transformation from correlation functions
\beq
\Delta_{ab} = \frac{\langle (e_a - \langle e_a\rangle)(e_b - \langle e_b \rangle)\rangle}{\sigma_a\sigma_b},
\eeq
where $\langle\ldots\rangle$ is the expectation value and $\sigma$ is the
standard deviation, we transform events into dimensionless variables:
\beq
\nu_a = \frac{x_a-\langle x_a\rangle}{\sigma_a},
\eeq
and form an adjacency matrix of path coefficients, or probabilities $p_{ab}$, such that:
\beq
\Delta_{ab} &=& \sum_c p_{bc}\,\Delta_{ac}.\\
       &=& \sum_c \alpha_i(\Psi_c).
\eeq
where $\Psi_c$ is a single directed acyclic path joining the source $x_a$ to $x_bb$.
The valuation function $\alpha_i()$ is none other than an  assessment function of agent $A_i$.
This linear relation can be inverted to yield weights on the edges of the semantic
graph, where the $p$ matrices act like probability amplitudes along the path, between the pure states:
\beq
p^T p = 1.
\eeq
Semantic spacetime is thus the graph in which the nodes are promise valuations
for semantic relationships of observable attributes. The valuations are
an undirected representation of the directed graphs, since we have factored
out the causal semantics by defining paths with components $\psi$.
The set of semantic spacetime nodes is now formed from interior variables to the
physical agents, and may be used to form a global picture of the world of the
agents for some `godlike' observer.

\bigskip
\subsection{Remarks about probability}

As a generic toolbox, probability is a powerful abstraction, albeit an
ambiguous one.  When invoking probability, we are implicitly asking:
how can we understand the process which counts data without all the
information (or how can we guess how data would be counted, if we
could do the experiment)? The probability is an incomplete
description. It eliminates important questions.  How did we
normalized? Over what spacetime scale are we averaging the frequencies
or computing the propensities? What accuracy can we expect?
It is up to the researcher to supplement these details.

\begin{figure}[ht]
\begin{center}
\includegraphics[width=8cm]{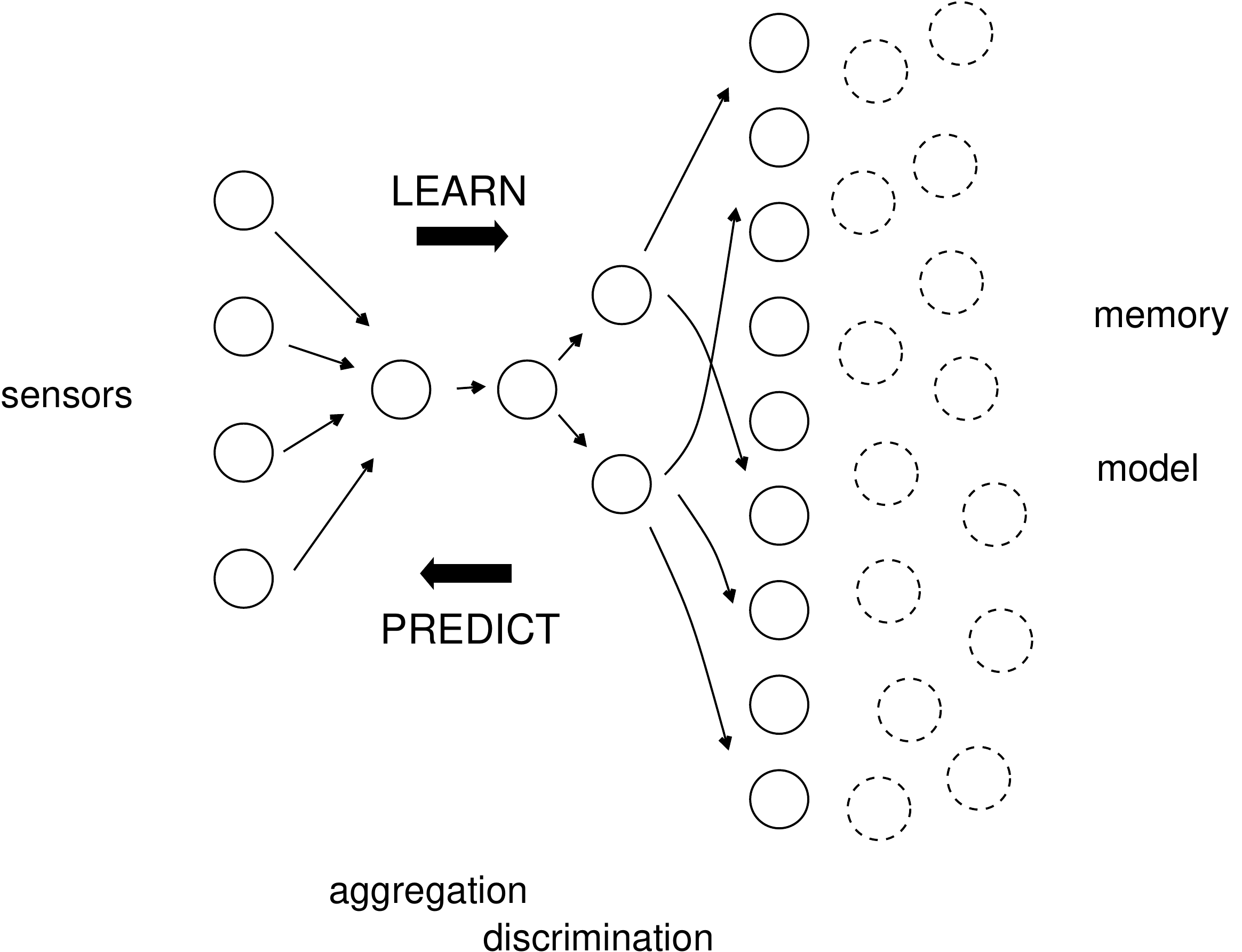}
\caption{\small From a parallel sensor array, through a pipeline of influence, to a
  memory representation. First, independent signals must be collated
  and classified, then stored in some interior representation, so
  first there is aggregation and then sparse representation. The final
  size is likely much larger than the original sensory
  image, so one must be economical in storage representation.\label{pipeline}}
\end{center}
\end{figure}
If we want to make a quantitative theory of promises, in terms of
either potentials or probabilities, then which variables should we
choose to count?
Some critical points:
\begin{itemize}
\item A successful model depends on how the pure state classes $\omega_a$ are chosen to parameterize behaviour. 
\item An event does not imply intent and an intention does not imply the generation of any events.
\item An event that appears to fulfil a promise, that was not given, is not evidence of actual intent, but if repeated enough times might be evidence of an effective intention, or an inherited bias.

\item Small scale changes should not infer large scale predictions.
\item Conditional promise is an indication of causal behaviour, but a
  conditional probability is only a {\em possible indication} of
  causal behaviour.  A joint event is a correlation; writing it as a
  pre-conditional expression is only a biased rescaling.
\end{itemize}
Finally, a promise that generates no associated events to fulfil its intention may be considered a lie.

\bigskip
\section{Quantum and statistical state descriptions}\label{densitymatrix}

Turning directly to probability may not always the most appropriate
representation for a physical system evolving in time. It is always
preferable to model the interior configurations of dynamical variables
directly if we know what they are.

From Quantum Mechanics, we learn an important lesson about how a
partial model can tell us much more than probability alone can, albeit
with certain caveats\cite{qm,Morgan_2022,WETTERICH20181}.
A vector (Hilbert)
space representation of allowed states for a dynamical process gives
us an algorithm for the feature axes, or pure states, as the solution
of a boundary problem as eigenvectors of the dynamical constraints.

A bounded quantum mechanical problem has some similarities to an autonomous agent. Each
bounded region contains its dynamical content autonomously.  We can
begin by assuming that such states are `within' the system, and
therefore we can think of it as an agent. For example, a hydrogen
atom, or a square well potential has a set of states that are
determined by the solution of a constraint equation that specifies a
spacetime geometry for a process.

\subsection{States generated by spacetime geometry}

A quantum state formalism leads to a quasi-statistical description,
whose interpretation is still much debated. It has some similarities
with Boltzmann's statistical mechanics\cite{feynmansm}: the two can be
connected into a unified treatment at a certain scale, at the expense
of overlooking a number phenomena that average out on a bulk scale.
Conversely, a system traditionally thought of as a fundamental can be
written as a statistical theory by re-framing it on a vector or
Hilbert space. Most of classical physics is unbounded and concerns
weakly constrained motion (of the first kind), which sets it apart
from agent systems in Promise Theory, which start from bound states.

Tradition in probabilistic methods encourages us to imagine a single universal passage of time
that we might call observer time, in which some clock on a wall tells
us its unambiguous count from some arbitrary event. In practice, this
is not a useful abstraction in a distributed system. Wherever
independent processes run in parallel, each agent has its own clock that
does not necessarily advance at the same rate as others'.
Quantum Mechanics also involves a number of different time semantics
that are left to the observant reader to interpret.
The interpretation of these states as generators for a probabilistic
interpretation would normally indicate the existence of a deeper level, faster process,
whose statistics are reflected in the probabilistic amplitudes.

\begin{figure}[ht]
\begin{center}
\includegraphics[width=8.5cm]{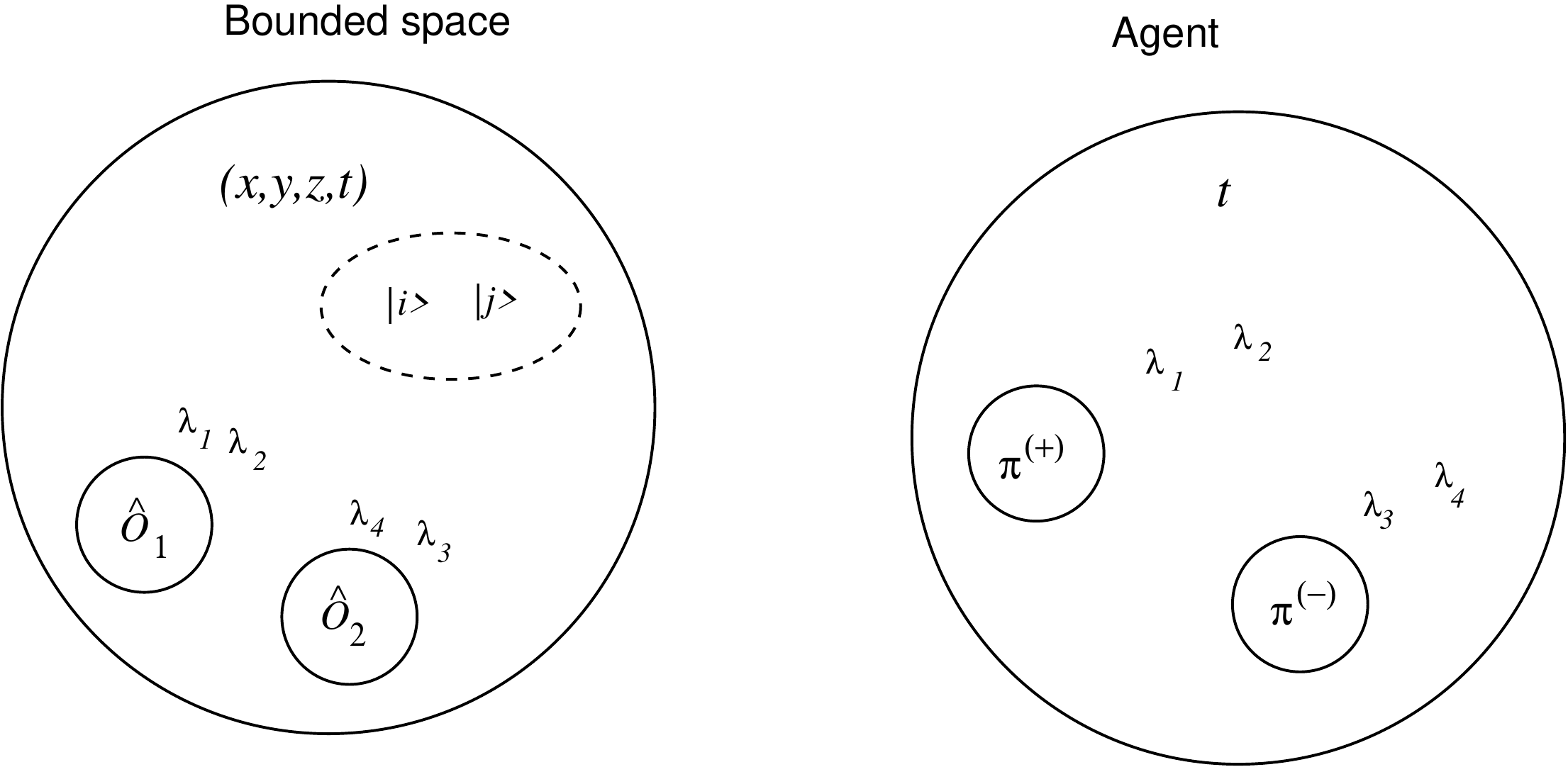}
\caption{\small Representing bounded capabilities. In quantum
  mechanics, the boundary of a geometry and potential function of a spacetime
  arrangement (e.g. a square well or isolated hydrogen atom) forms a constraint on an energy distribution process,
  leading to a set of supportable eigenstates. Observables
  representing differential physical processes each have a set of
  supportable eigenstates that give observed value of a single
  observation. Agents are independent bounded regions, each of which
  has interior resources and states, outside of normal spacetime.
  These are not represented by differential geometric equations, but
  rather by computable functions. The representation of allowed states
  is analogous in both cases.\label{interior}}
\end{center}
\end{figure}

In Quantum Mechanics, a system of study has a spacetime geometry
which, ideologically, precedes everything else.  One begins by making
these boundary conditions consistent with continuity and conversation
laws. The dynamics of a wave vector are determined by these factors.
Observables correspond to spacetime `processes' that are encoded as
differential (or matrix) operations, with appropriate dimensions and
geometry.

The shift to a conceptual space of possible states, formed by linear combinations of
continuous eigenfunctions,
\beq
\hat O |i\rangle = \lambda^{O}_i\, |i\rangle.
\eeq
takes a step towards a partly-knowable interior dynamics.
We can map this picture onto autonomous agents as an alternative way of approaching
the integration of independent agents\cite{virtualmotion1}. 
Each isolated system can be associated with a density matrix
is assumed to have a complete set of pure states $\i\rangle \leftrightarrow \omega_i$,
which spans and plausibly characterizes its
instantaneous activity. A complete representation, assumes that one knows
its interior generating function $H$. One uses the terminology of `pure state' to mean
one that is an eigenstate of its constraints (represented by the Hamiltonian), so that a
linear combination is an interior superposition, simultaneous with respect to interior time.
\beq
\psi_\text{pure}(x,t) = \sum_i \; c_i(x,t)\,|i\rdirac, ~~~~~~(\sum_i |c_i|^ 2 = 1)
\eeq
The spherical constraint (referring to the surface of th Bloch sphere)
indicates that these pure states are specially constrained
by a rigid evolution, in which the system moves from pure state to pure state,
rotated only by a phase (like a gauge choice), thus conserving the
instantaneous probability.

A mixed state is a more arbitrary convex combination of the same complete set, in which
the distribution of probabilities is not dynamically constrained on the inside of the Bloch sphere:
the weighting is assumed to be assembled from an ensemble or statistical mixture.
Thus it now refers to longer coarse grains of shared co-time between agents.
The definition of time has changed, since the smallest
interval is now the time average of many pure configurations\footnote{It is commonly stated that a classical system cannot be in a
superposition of pure states, only as an ensemble or mixed state. This
is not strictly true, as it depends on an assumed representation. A
boundary for what constitutes the interior of the system as opposed to
what represents multiple copies or trials, and a separation of
timescales can easily give the impression of a system being in
multiple equivalent states.  Superpositions of interior states are
possible. For example, a fish may be light and dark in patches when
examined in detail. When measured, the net average may still collapse
into either light or dark as a function of the measurement process.}.
\beq
\psi_\text{pure}(\overline x,\overline t) = \sum_i \; m_i(\overline x,\overline t)\,|i\rdirac, ~~~~~~(\sum_i m_i = 1)
\eeq
The same temporal redefinition applies when autonomous agents are joined into a
statistical ensemble.

\subsection{Density matrix representation for agent interactions}

The density matrix is often introduced to relate states and probabilities of interconnected systems\cite{feynmansm}.
One can treat a chain of agents as a chain of interacting pseudo-quantum
systems too.

Along each hop of an agent pipeline, or path joining agents in a
shared process, we can also represent the state evolution from the
perspective of either sender or receiver, using a density
matrix\cite{qm}.  From the perspective of autonomous agents, at any
scale, the picture is one where a local process in one agent observes
another remote process in a neighbouring agent, and we can view the
interaction as a temporary composite system.

\begin{figure}[ht]
\begin{center}
\includegraphics[width=9cm]{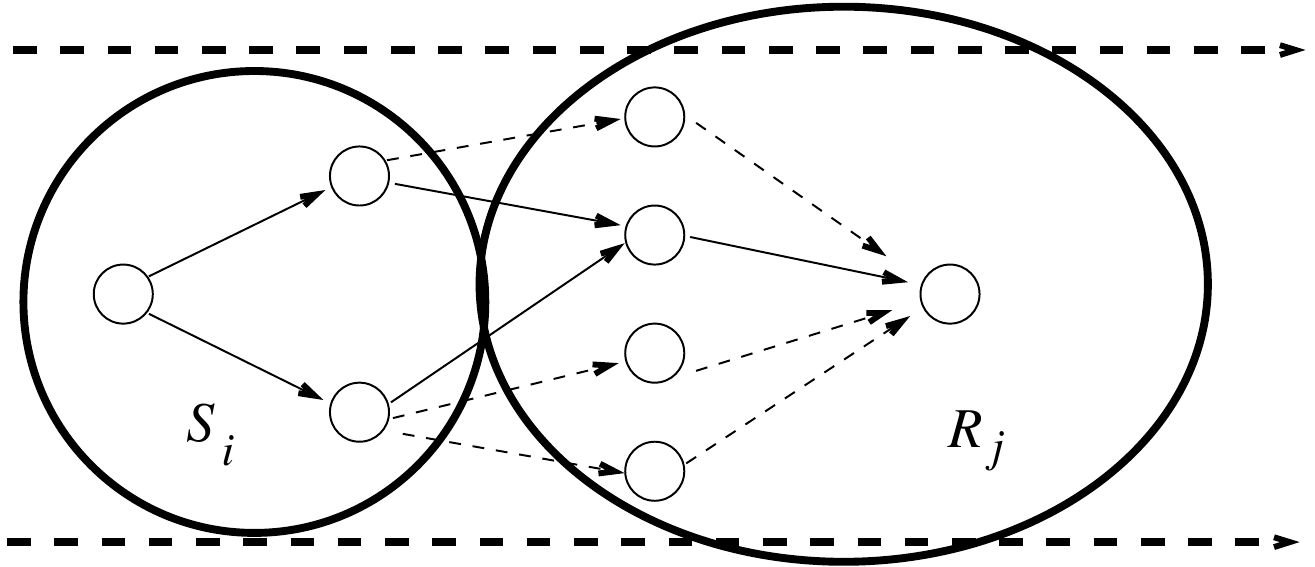}
\caption{\small Motion along a channel from coarse-grained superagent to
superagent provides the most general scaled picture of transitions. This
view leads naturally to a density matrix formulation due to the separation
of interior and exterior process variables. The scale at which we
describe this process now matters. If the circles represent a scale $S$, then the
arrows are shown at scale $S-1$.\label{coarse}}
\end{center}
\end{figure}

The density matrix for a local agent lives on the same variable
space as the evolving interior processes, but contains information from the
neighbour-driven mixture.
In a privileged observer view, 
the space and time variables are exterior variables that are common to
the composite system. The density operator on these states $\rho$ is an operator on $A_i$,
which has interior states $I_i$. The exterior states belong to another co-entangled
system $A_j$. It's convenient to relabel these as sender and receiver along an oriented
path $S_i$ and $R_j$ (figure \ref{coarse}).

The implication is that there are two timescales: a fast interior process which is
being observed over a partially equilibrated slowly-moving macrostate.
Macrostates are usually thought of as macroscopic systems, but that's relative.

To foster associations between agents and quantum states, we introduce a Dirac notation for the states
of promise agents in semantic spacetime. The double bra-ket lines
should remind us that the nature of these states is not to be
associated necessarily with a quantum Hilbert space, indeed need not be defined in detail here,
except to say that they are assumed normalizable. Depending on the
nature of the agents a complete set of states spanning the agent's
`possibility space' may take various forms.  Let 
\beq 
| A_i\rdirac
\eeq 
be the interior states of agent $A_i$, which are unobservable to
other agents.

Consider a capacity allocation process $\psi(A_i, A_j)$. The joint distribution
of available resources can be written:
\beq
\psi(S_i,R_j) = \sum_i c_i(R_j)\, I(S_i),\label{joint}
\eeq
for some set of interior states $I_i$ and exterior states $E_j$. 
\beq
I_i(S_i) &=& \ldirac S_i | I_i\rdirac = \ldirac S_i | i\rdirac\\
E_j (R_j) &=& \ldirac R_j | E_j\rdirac = \ldirac R_j | j\rdirac
\eeq
The most general interaction distribution would be a
product state\cite{virtualmotion1}:
\beq
\psi \sim S\otimes R \sim I \otimes E,
\eeq
which implies an accepted channel for promise keeping $\psi$, with a complete set of states
in the sample space $\Omega$, denoted $|I_i\rdirac$ and $|E_k\rdirac$:
\beq
|\psi\rdirac = \sum_{i,j} \; c_{ij}\;|I_i\rdirac |E_k\rdirac.
\eeq
In Dirac notation, (\ref{joint}) becomes 
\beq
\psi(A_i,A_j) &=& \ldirac A_j | \ldirac A_i | \psi\rdirac\\
              &=& \sum_{i,j} c_{ij} \, \ldirac A_i|I_i\rdirac \ldirac A_j | E_j\rdirac,
\eeq
and comparing to (\ref{joint}), we have the coefficients expressed as projected linear combinations of
exterior process availability:
\beq
c_i(A_j) = \sum_j c_{ij} \; \ldirac A_j | E_j \rdirac.
\eeq
So, if we consider this simple two-agent promised transition
process $\pi^{-}\pi^{+}$ to be an operation on these states, without
needing to explain its representation, then we can express this
transition as an effective current:
\beq
\pi^{(-)}_{\delta q}\pi^{(+)}_{\delta q} = \sum_{i,j}\; | I_i\rdirac |E_j\rdirac \cdot \ldirac E_j | \ldirac I_i|,
\eeq
so that the expectation value, which corresponds to the availability of a transition with operator
\beq
T_{q} \equiv  \pi^{(-)}_{\delta q}\pi^{(+)}_{\delta q}
\eeq
is
\beq
\ldirac T_{q} \rdirac &=& 
\sum_{i,j, i',j'} c_{ij}^{(-)}  c_{i'j'}^{(+)}
\ldirac E_j | \ldirac I_i|,
\pi^{(-)}_{\delta q}\pi^{(+)}_{\delta q}
|I_{i'}\rdirac | E_{j'} \rdirac \\
&=& \sum_{i,i'} \; \rho_{ii'} \; \ldirac I_i|\;T_{q}\,|I_{i'} \rdirac\\
&=& \text{\rm Tr}\; (\rho \; T_{q})
\eeq
for density matrix $\rho$, which can be written as a convex combination of projections for the pure states,
each with availabilities or weighting $w_i$:
\beq
\rho &=& \sum_i \; w_i \;|i\rdirac \ldirac i||, ~~~\text{where} ~~~\sum_i w_i = 1,\\
w_i &=& \sum_{j}\; c_{ij}^{(-)}c_{ij}^{(+)},
\eeq
and $w_i$ is the relative availability for a channel capacity reservation.
\begin{itemize}
\item A pure state $\psi_i = \sum_i\; w_i\; |i\rdirac$, where $\sum_i w_i^2 = 1$. The location of $q$ can be rotated from one location to the next by a phase.

\item A mixed state is  $\psi_i = \sum_i\; w_i\; |i\rdirac$, where $\sum_i w_i = 1$ the location of $q$ is neither in one location or another.
\end{itemize}
The off-diagonal elements express the entanglements of neighbouring
agents that establish channels for transmission $\Delta q$.  For a
so-called pure state, only one weight affinity $w_i$ is non-zero.

The main purpose of writing agents in this way is to make contact with
quantum methods, which appear at least superficially similar to that of autonomous agents.
I'll leave this identification here for others to pursue.

\section{Conclusions}

What should we expect of a quantitative description of agents shaped
by their promises?  In scientific studies, we are usually trying to
construct an accurate representations of phenomena: to understand the
key players in a process, rates of change, and measurable quantities,
etc.  In simple studies of a single variable, elementary statistics
could be sufficient to describe the phenomenon, because we can
incorporate all assumptions as an introduction to the study.  In
fields like immunology and neuroscience, where one is trying to model
the complex interactions between very different kinds of cells,
i.e. with different semantics, semantics and dynamics do not separate
cleanly.

In this review, I've tried to show how one may get started in
formulating problems so as to be compatible with promise theoretic principles.
The precise mixture of dynamics and semantics involved in any
explanation prioritizes these matters differently. Principally
quantitative theories (e.g. physics) are built on a long history of
generally accepted assumptions that we call semantics, and therefore they
tend to be weak in encoding these fine points formally. Deeply semantic theories, like the
biological sciences, may struggle to converge on a set of suitable
idealized approximations that permit a quantitative result to be defined.

Recently, knowledge acquisition, generative reasoning, and decision theory have been the main drivers of
modelling, for data associated with `knowledge capture', `machine learning' and
`artificial intelligence'. In all such representations, one is looking
for an approximation: a threshold for choosing one decision pathway
over another.  No matter how one dresses up a
threshold, by calculation, it must remain arbitrary and
thus speculative. The principle of convergence to a stable end state is one of the
few motives for selecting without such a threshold.

There are two dominant ideas about reasoning in science
today\footnote{The hybrid Tsetlin automaton model of voting bridges
  this divide from the bottom-up\cite{Granmo2018TheTM}.}:
\begin{itemize}
\item Logical reasoning based on symbolic propositions that are
  semantically fragile and complex to establish.

\item Bayesian probabilities that may be robust to establish, but are computationally expensive to
  update and potentially ambiguous.

\end{itemize}
These are not the only possibilities, yet
increasingly researchers go straight for probability as a descriptive
tool, and waive the details of a causal deterministic
process. Probability incorporates uncertainty, so it feels like an
attractive lingua franca for theorists; however, relying on statistics
alone potentially conflates matters of data collection with
explanation. Adopting the certainty of logic, one may have to resort to dishonesty about
classifications and beliefs. Adopting probabilistic forms, one risks merely
replacing one form of dishonesty with another, where the use of probability itself is
what stands in the way of a deeper understanding.

Promise Theory and its agent dynamics offer a possible supplement to
balance these, involving a composition of processes from several
weakly coupled or mostly independent parts. It also relates to models
of swarm intelligence and cellular automata\cite{swarm,swarm2,swarmmodels,kazadithesis}.
From the perspective of Promise Theory, reasoning is about the
formation of storylines, or pathways through representative nodes in a
semantic spacetime, and intent is about aligning with the outcomes of
those story paths.

Promise Theory's domain of quantitative application involves many
forms of distributed cooperative behaviour: resilient architectures
with redundant dependencies; fidelity of distrubuted configurations,
coherent cooperation, and decentralized team performance. Semantic
tasks include defining the shared purposes, knowledge graphs,
authority, etc.  The modelling of monetary networks is a natural
promise theoretic task, indeed all forms of transport network
cooperation and throughput. Modelling the value of workers in
collaborative enterprises, whether human or robot. It has much to
contribute to the mechanics of complex systems in particular. It's
chief disadvantage currently lies in being unfamiliar.

In probability, the question of normalization and its semantics is
always a persistent headache, especially when a system is time
varying. This becomes even worse for autonomous agents, which have no
controller for learning and managing the population from an external
`Galilean' perspective.  For promise theoretic studies, the concern is
that agents, which can reproduce the kinds of elementary phenomena we
observe in nature, must have abundant interior resources in order to
represent the processes of the natural world. There must be `plenty of
room at the bottom'\cite{feynmanbottom} in order to realize phenomena
as autonomous agents---all of which begs the question: how does nature really
do it?

\bibliographystyle{unsrt}
\bibliography{spacetime,bib}

\end{document}